%% file: main.tex
\documentclass[10pt,twocolumn,letterpaper]{article}

\usepackage[pagenumbers]{iccv} 


\usepackage{bbm}
\usepackage{amsthm}
\usepackage{booktabs}
\usepackage{algorithm}
\usepackage{amssymb}
\usepackage{multirow}
\usepackage{wrapfig}
\usepackage{makecell}
\usepackage{colortbl}
\usepackage{bbding} 
\usepackage{color}
\usepackage{verbatim}
\usepackage{pifont} 
\newtheorem{theorem}{Theorem}[section]

\newtheorem{lemma}[theorem]{Lemma}

\newtheorem{definition}[theorem]{Definition}

\usepackage{setspace}
\usepackage{algorithmicx}
\usepackage{amssymb}
\makeatletter

\newcommand{\Rmnum}[1]{\expandafter\@slowromancap\romannumeral #1@}
\makeatother

\definecolor{C0}{rgb}{0.121569, 0.466667, 0.705882}
\definecolor{C1}{rgb}{1.000000, 0.498039, 0.054902}
\definecolor{C2}{rgb}{0.172549, 0.627451, 0.172549}
\definecolor{C3}{rgb}{0.839216, 0.152941, 0.156863}
\definecolor{C4}{rgb}{0.580392, 0.403922, 0.741176}
\definecolor{C5}{rgb}{0.549020, 0.337255, 0.294118}
\definecolor{C6}{rgb}{0.890196, 0.466667, 0.760784}
\definecolor{C7}{rgb}{0.498039, 0.498039, 0.498039}
\definecolor{C8}{rgb}{0.737255, 0.741176, 0.133333}
\definecolor{C9}{rgb}{0.090196, 0.745098, 0.811765}
\definecolor{iccvblue}{rgb}{0.21,0.49,0.74}
\usepackage[pagebackref,breaklinks,colorlinks,allcolors=iccvblue]{hyperref}
\usepackage{xcolor}
\definecolor{aliceblue}{RGB}{176,223,229}
\newcommand{\CC}{\cellcolor{aliceblue}}

\newcommand{\imineq}[2]{\vcenter{\hbox{\includegraphics[height=#2ex]{#1}}}}

\def\paperID{9779} 
\def\confName{ICCV}
\def\confYear{2025}

\title{\texttt{CoRe$^2$}: \textit{Collect, Reflect and Refine} to Generate Better and Faster}

\author{%
  Shitong Shao$^{1}$,\space\space\space
  Zikai Zhou$^{1}$,\space\space\space
  Dian Xie$^{1}$,\space\space\space
  Yuetong Fang$^{1}$,\space\space\space
  Tian Ye$^{1}$,\space\space\space
  Lichen Bai$^{1}$,\space\space\space
  Zeke Xie$^{1}$\thanks{Corresponding author.} \\
  $^1$The Hong Kong University of Science and Technology~(GuangZhou)\\
  \scriptsize{\texttt{\{sshao213, zikaizhou, dianxie, yfang870, tye610, lichenbai, zekexie\}@hkust-gz.edu.cn; *: Corresponding author
  }}\\
}

\begin{document}
\twocolumn[{%
\renewcommand\twocolumn[1][]{#1}%
\maketitle\begin{center}\centering\includegraphics[width=1.0\linewidth]{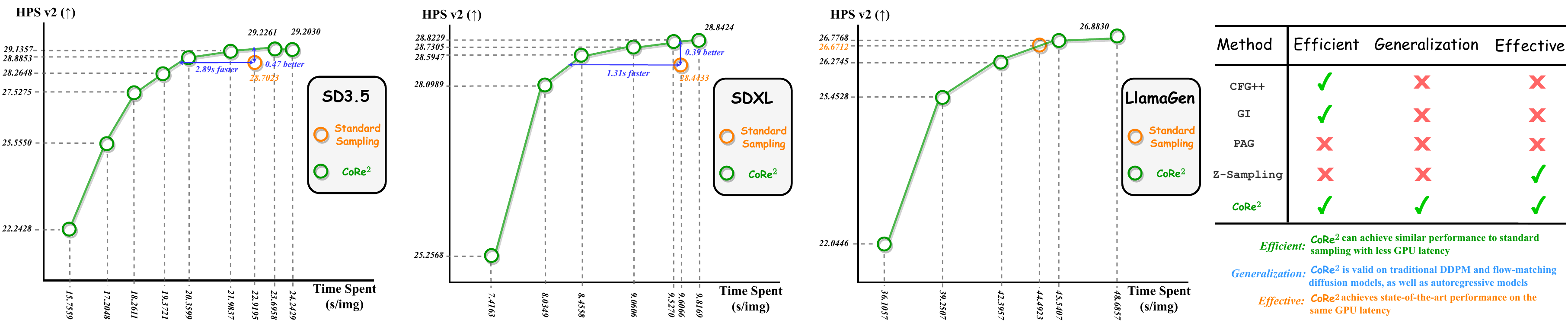}
\captionof{figure}{\textbf{Left:} Our proposed \texttt{CoRe}$^2$ achieves an excellent balance between performance and efficiency across SD3.5, SDXL, and LlamaGen. Specifically, for SD3.5 and SDXL, it produces more faithful, realistic, and detailed images with high semantic consistency, while significantly reducing computational overhead compared to standard sampling. Even on LlamaGen, \texttt{CoRe}$^2$ enhances the model's generative capabilities with only a minimal increase in computational cost. \textbf{Right:} Compared to previous inference-enhanced algorithms, \texttt{CoRe}$^2$ achieves optimal performance across the three dimensions of efficiency, generalization, and effectiveness.}
\label{fig:first_presentation}
\end{center}%
}]
\maketitle
\input{sec/0_abstract}    
\input{sec/1_intro}
\input{sec/2_preliminary}

\input{sec/3_method}
\input{sec/4_experiment}

\vspace{-5pt}
\section{Conclusion}
\vspace{-3pt}
This paper presents a groundbreaking inference mechanism named \texttt{CoRe}$^2$, which seamlessly integrates efficiency, effectiveness, and generalization. It stands out as the first plug-and-play inference-enhanced algorithm to demonstrate simultaneous effectiveness across both DM and ARM. The core principle of \texttt{CoRe}$^2$ revolves around gathering data to train a weak model that reflects ``easy-to-learn'' aspects, followed by the application of W2S guidance to refine ``difficult-to-learn'' components, thereby significantly enhancing the high-frequency details in images.

{
    \small
    \bibliographystyle{ieeenat_fullname}
    \bibliography{main}
}

\input{sec/appendix}

\end{document}

%% file: sec/0_abstract.tex
\begin{abstract}
Making text-to-image (T2I) generative model sample both fast and well represents a promising research direction. Previous studies have typically focused on either enhancing the visual quality of synthesized images at the expense of sampling efficiency or dramatically accelerating sampling without improving the base model’s generative capacity. Moreover, nearly all inference methods have not been able to ensure stable performance simultaneously on both diffusion models (DMs) and visual autoregressive models (ARMs). In this paper, we introduce a novel plug-and-play inference paradigm, \textbf{\texttt{CoRe}$^2$}, which comprises three subprocesses: \textit{\textbf{Co}llect}, \textit{\textbf{Re}flect}, and \textit{\textbf{Re}fine}. \texttt{CoRe}$^2$ first collects classifier-free guidance (CFG) trajectories, and then use collected data to train a weak model that reflects the easy-to-learn contents while reducing number of function evaluations during inference by half. Subsequently, \texttt{CoRe}$^2$ employs weak-to-strong guidance to refine the conditional output, thereby improving the model's capacity to generate high-frequency and realistic content, which is difficult for the base model to capture. To the best of our knowledge, \texttt{CoRe}$^2$ is the first to demonstrate both efficiency and effectiveness across a wide range of DMs, including SDXL, SD3.5, and FLUX, as well as ARMs like LlamaGen. It has exhibited significant performance improvements on HPD v2, Pick-of-Pic, Drawbench, GenEval, and T2I-Compbench. Furthermore, \texttt{CoRe}$^2$ can be seamlessly integrated with the state-of-the-art Z-Sampling, outperforming it by 0.3 and 0.16 on PickScore and AES, while achieving 5.64s time saving using SD3.5. \textcolor{red}{Code is released at \href{https://github.com/xie-lab-ml/CoRe/tree/main}{\textcolor{red}{\texttt{CoRe}$^2$}}}.
\vspace{-6pt}
\end{abstract}

%% file: sec/1_intro.tex
\section{Introduction}
\label{sec:intro}\begin{figure}[t]
    \centering
    \includegraphics[width=1.0\linewidth]{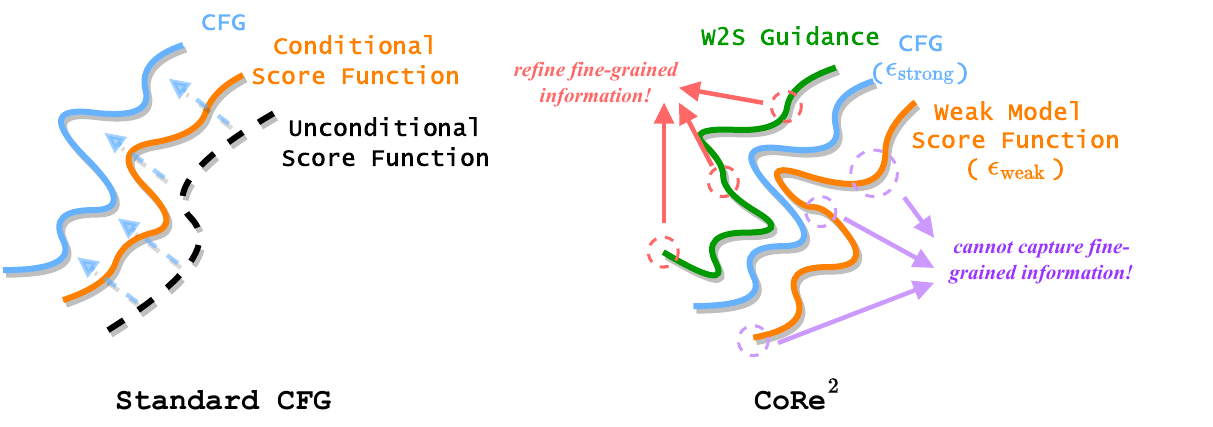}
    \vspace{-22pt}
    \caption{A concise explanation of why \texttt{CoRe}$^2$ is effective (diffusion model for an example): we train a weak model to reflect the easy-to-learn components. Then, W2S guidance is employed to refine the more (fine-grained) difficult-to-learn components.}
    \label{fig:empirical_vis_core2}
    \vspace{-14pt}
\end{figure}\begin{figure*}[t]
    \centering
    \includegraphics[width=0.95\linewidth]{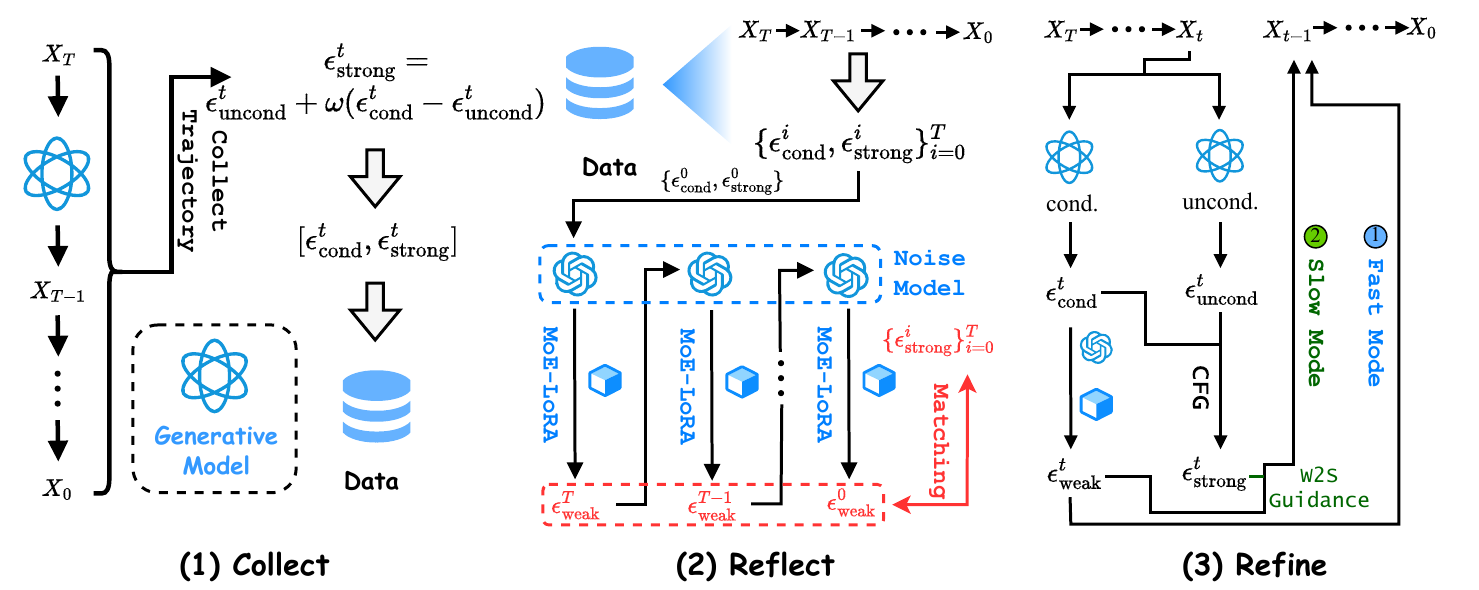}
    \vspace{-12pt}
    \caption{Overview of \textit{\textbf{Co}llect}, \textit{\textbf{Re}flect} and \textit{\textbf{Re}fine} (\textbf{\texttt{CoRe}$^2$}). We initially generate trajectories corresponding to CFG to collect data. Next, we train a weak model (i.e., the noise model equipped with MoE-LoRA) to capture the mapping from the conditional output to the CFG output, reflecting the easy-to-learn content. Finally, we employ W2S guidance to refine the conditional output (i.e., the fast mode) and the CFG output (i.e., slow mode), thereby enhancing the critical fine-grained information that is challenging to learn.}
    \label{fig:total_framework_core2}
        \vspace{-12pt}
\end{figure*}
Over the past few years, text-to-image (T2I) diffusion models (DMs)~\citep{ddpm_begin,ddim,sde,dpm_solver,controlnet,zigzag,shao2023catch,shao2025ivmixed} and visual autoregressive models (ARMs)~\citep{llamagen,muse} have garnered significant attention within the domain of generative visual intelligence due to their ability to produce high-quality images. Compared to strengthening the generating capabilities of generative models during the pre-training phase, considering post-training or even training-free strategies directly during the inference stage requires less overhead to improve the fidelity of synthesized images~\citep{golden_noise,chung2024cfg++,guidance_interval,pag_diffusion,noiserefine}. Through the implementation of inference-enhanced algorithms, such as classifier-free guidance (CFG)~\citep{nips2021_classifier_free_guidance}, CFG++~\citep{chung2024cfg++}, Guidance Interval~\citep{guidance_interval}, PAG~\citep{pag_diffusion}, and Z-Sampling~\citep{zigzag}, the sampling process of DMs has achieved significant advancements.

However, prior solutions predominantly excel either by enhancing visual quality at the cost of significant additional inference latency~\citep{zigzag,cvpr22_kd_guided,liu2024correcting} or by ensuring efficiency but falling short in terms of performance~\citep{chung2024cfg++,ddim,guidance_interval}. These algorithms fail to allow generative models to achieve quick and high-quality sampling. In addition to this, these algorithms are primarily designed for DMs, with their usability even more narrowly confined to specific architectures like SD1.5~\citep{SDV1.5}, SD2.1~\citep{SD2.1} and SDXL~\citep{SDXL}. Their performance falls short on large-scale, flow-matching-based foundation models, such as SD3.5~\citep{SD35} and FLUX~\citep{FLUX}. Meanwhile, several methods, like PAG~\citep{pag_diffusion} and Z-Sampling~\citep{zigzag}, which are architecture-specific and depend on inversion operations unique to DMs, making them incapable of generalizing to models like ARMs. To address the current limitations of inference-enhanced algorithms, which struggle to simultaneously ensure effectiveness, efficiency, and generalization (see Fig.~\ref{fig:first_presentation}, right), we propose a novel and plug-and-play sampling framework named \textbf{\texttt{CoRe}$^2$}, which consists of three stages: \textit{\textbf{Co}llect}, \textit{\textbf{Re}flect}, and \textit{\textbf{Re}fine}. This paradigm neither relies on the \textit{forward} and \textit{reverse} processes of DM nor is it architecture-specific, and it achieves self-improvement (see Fig.~\ref{fig:first_presentation}, left) on both DM and ARM without the need for an external assistant, a reward model and a new dataset.


Specifically, \texttt{CoRe}$^2$ relies exclusively on CFG, a technique that enhances the coherence between image and text for both DM and ARM. During the \textit{Collect} phase, it collects the sampling trajectories of the generative model to construct a mapping from the conditional output to the CFG output. Unlike conventional CFG distillation~\citep{iclr22_progressive,icml23_consistency,sid_lsg}, \texttt{CoRe}$^2$ can pre-store the data, thereby eliminating the need to load the redundant pre-trained generative model into GPU memory during the \textit{Reflect} stage. In the \textit{Reflect} phase, we exclusively rely on a weak model with limited capacity to learn the mapping from the conditional output to the CFG output, thereby reflecting the easy-to-learn content within CFG. In practical experiments, we implement this weak model using a noise model~\citep{golden_noise,noiserefine}. This way exhibits the following merits: \textbf{(1)} Noise model is extremely lightweight, incurring only a 2.18\% additional GPU latency compared to CFG sampling on SDXL; \textbf{(2)} Its limited fitting capacity suits our goal of reflecting only the easy-to-learn content in CFG, poorly capturing the difficult-to-learn aspects; \textbf{(3)} Its optimization objective is similar to CFG distillation, allowing it to replace CFG at suitable sampling intervals with minimal performance impact, thereby accelerating the sampling process (see the fast mode in Fig.~\ref{fig:total_framework_core2}).

In the final stage, we design two distinct modes for inference: the fast mode and the slow mode, tailored for the later and earlier sampling intervals. The fast mode operates by directly utilizing the well-trained noise model to refine the conditional output, efficiently emulating the CFG output. In contrast, the slow mode adopts a more nuanced approach by treating the CFG as the strong model and the noise model as the weak model. Through weak-to-strong (W2S) guidance~\citep{bad_version_diffusion}, \texttt{CoRe}$^2$ works to enhance the generative model’s overall performance, improving both visual quality and semantic faithfulness. By carefully balancing the utilization frequencies of the slow and fast modes, we can generate higher-quality images while reducing GPU latency compared to the standard sampling.


\noindent{Our contribution consists of the following three parts:}
\vspace{-1pt}
\begin{enumerate}[label=$\bullet$] 
    \item We propose the novel, plug-and-play \texttt{CoRe}$^2$ that decouples from specific generative model paradigms while delivering excellent performance in terms of three dimensions: effectiveness, efficiency, and generalization.
    \item We empirically demonstrate the mechanism that our approach excels at refining intricate, high-frequency details that are often challenging for models to learn. We also further provide a rigorous theoretical proof to explain why \texttt{CoRe}$^2$ achieves such remarkable effectiveness.
    \item Our experimental results demonstrate that \texttt{CoRe}$^2$ surpasses the vast majority of state-of-the-art (SOTA) methods across various DMs, including SDXL, SD3.5, and FLUX (see Appendix~\ref{apd:flux_experiment}), as well as ARM such as LlamaGen within renowned benchmarks such as HPD v2, Pick-of-Pic, Drawbench, GenEval, and T2I-Compbench.
\end{enumerate}

%% file: sec/2_preliminary.tex
\section{Preliminaries}
\label{sec:preliminaries}

\noindent{\bf Diffusion Model and Autoregressive Model.} Both DM and ARM utilizes a multi-step denoising paradigm to iteratively generate high-quality images during the inference process. Starting with a latent variable ${x}_{T}$, this variable may adhere to a Gaussian distribution (in the context of DMs~\citep{ddpm_begin}) or consist of masked tokens (in the case of ARMs~\citep{llamagen}). These methods fundamentally aim to approximate the (abstract) estimator $p({x}_{t-1}|{x}_{t})$ ($1 \leq t \leq T$) during the training phase, thereby enabling sequential sampling in the form of $p({x}_{0}|\prod_{i=1}^{T}{x}_{i})=p(x_T)\prod_{i=1}^{T}p({x}_{t-1}|x_t)$ during inference, where $T$ denotes the number of sampling steps. It is important to note that $p(x_{t-1}|x_{t})$ can take on different forms depending on the model’s focus. In DMs, it corresponds to the prediction of the score function, while in ARMs, it refers to the prediction of a token. Within DMs, the encoder-only Transformer architecture is often employed to predict either the entire score function or the full tokens. This allows the sampling process to be refined as $p(x_{t-1}|x_t)=\int_{x_0} p(x_{t-1}|x_t,x_0)p(x_0|x_t)dx_0$, offering a structured and comprehensive approach to image generation. Regardless of whether it is an ARM or a DM, both can leverage CFG to enhance their generative capabilities. For the sake of simplicity, we employ the noise prediction notation commonly used in DMs to illustrate this process:\begin{equation}
    \begin{split}
        & \epsilon^t_\textrm{strong} = \epsilon_\textrm{uncond}^t + \omega(\epsilon_\textrm{cond}^t - \epsilon_\textrm{uncond}^t),\\
    \end{split}
\end{equation}
where $\omega$ refer to the CFG scale, which controls the strength of the guidance, while $t$, $\epsilon_\textrm{cond}^t$ and $\epsilon_\textrm{uncond}^t$ indicate the specific time step during the sampling process, the conditional output and the unconditional output, respectively.

\noindent{\bf CFG Distillation.} CFG enhances the fidelity of the generated results by performing an additional number of function evaluations (NFEs) under dropout-conditioning. Thus, injecting CFG's capability directly into the generative model can halve NFEs required during inference, thereby facilitating the deployment. This distillation paradigm was first introduced by~\citep{cvpr22_kd_guided}, and its core approach is to minimize the DM loss:
\begin{equation}
    \begin{split}
        & \mathcal{L}_\textrm{base} = \mathbb{E}_{(\epsilon,x_t,c)}\Vert\epsilon_\theta(x_t,t,c) - \epsilon \Vert_2^2,\\
    \end{split}
\end{equation}
while simultaneously minimizing the distillation loss:
\begin{equation}
    \begin{split}
        & \mathcal{L}_\textrm{distill} = \mathbb{E}_{(\epsilon,x_t,c)}\Vert\epsilon_\theta(x_t,t,c) - \epsilon_\textrm{strong}^t \Vert_2^2,\\
    \end{split}
\end{equation}
where $\epsilon_\theta$, $\epsilon$, $c$ and $\epsilon_\textrm{strong}^t$ stand for the score function estimator, the Gaussian noise, the text prompt and the CFG output. However, this distillation paradigm imposes a significant training burden—requiring both models that generate $\epsilon_\theta(x_t,t,c)$ and $\epsilon_\textrm{strong}^t$ to be loaded into GPU memory—and it fails to focus solely on the easy-to-learn aspects.

\noindent{\bf Noise Model.} Noise model was primarily introduced by Golden Noise~\citep{golden_noise} and is used to refine the initial Gaussian noise via a ViT backbone~\citep{VIT}, thereby enhancing both the fidelity and aesthetics of the synthesized images. Noise Refine~\citep{noiserefine} employs this strategy by performing CFG distillation in the first sampling phase. Unfortunately, its effect is limited and cannot be extended across the complete sampling trajectory. To solve this, we present a novel noise model architecture equipped with MM-DiT-Block~\citep{FLUX} and use MoE-LoRA~\citep{moelora} to achieve CFG distillation.

%% file: sec/3_method.tex
\section{\texttt{CoRe}$^2$ Framework}
\label{sec:method}

As illustrated in Fig.~\ref{fig:total_framework_core2}, \texttt{CoRe}$^2$ is implemented via a tripartite process, which is a highly generalized, enhanced inference procedure. As presented in Fig.~\ref{fig:empirical_vis_core2}, our algorithm essentially works by enhancing those difficult-to-learn yet crucial details of the model during the inference phase.

\vspace{-7pt}\paragraph{Stage-\Rmnum{1} \textit{Collect} $\left(\imineq{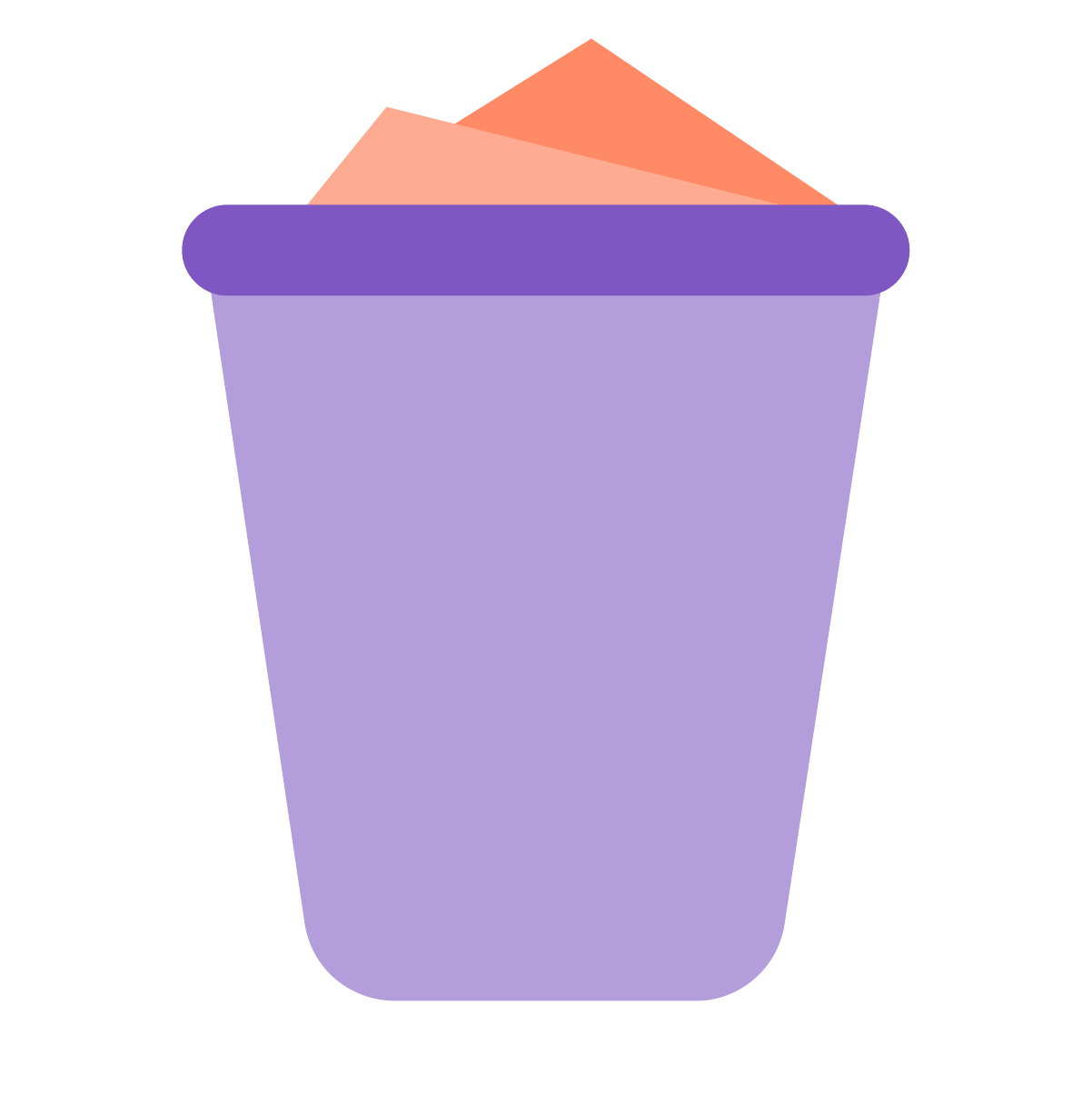}{2.4}\right)$.} This stage focuses on assembling a dataset precisely crafted for \textit{Reflect}. The collected data inherently comprise elements that are easy-to-learn, as well as components that are significantly difficult-to-learn. We have opted for CFG, an inference-enhanced technique widely used in DMs and ARMs, to collect the required data. For a model that requires $T$ sampling steps, the collected data for a single sample can be represented as $\{(\epsilon_{\textrm{cond}}^t, \epsilon_{\textrm{uncond}}^t, \mathcal{C}^t)\}_{t=1}^T$, where $\mathcal{C}$ stands for other conditions such as text embedding $G_\textrm{text}$ and timestep $t$. Due to the large size of $G_\textrm{text}$, storing them consumes significant disk space. To address this issue, we apply singular value decomposition (SVD) $f_\texttt{SVD}$ to decompose and reduce the dimensionality of the text embeddings prior to storage:\begin{equation}
    \begin{split}
        & U \times \Sigma \times V^T = f_\texttt{SVD}(G_\textrm{text}),\\
        \text{store}(&U[:,:\text{rank}],\Sigma[:\text{rank},:\text{rank}],V[:,:\text{rank}]). \\
    \end{split}
    \label{eq:svd}
\end{equation}
Here, $\Sigma$ denotes the diagonal matrix and rank indicates the reduced rank after dimensionality reduction. In our implementation, rank is set to 64 in SD3.5, SDXL and LlamaGen. Through this pipeline, we obtain the synthesized dataset $\mathcal{D} = \{ (\epsilon_{\textrm{cond}}^{(t,i)}, \epsilon_{\textrm{uncond}}^{(t,i)}, \mathcal{C}^{(t,i)}) \}_{(t=1,i=1)}^{(T,N)}$.

\begin{figure}
    \centering
    \includegraphics[width=1.0\linewidth]{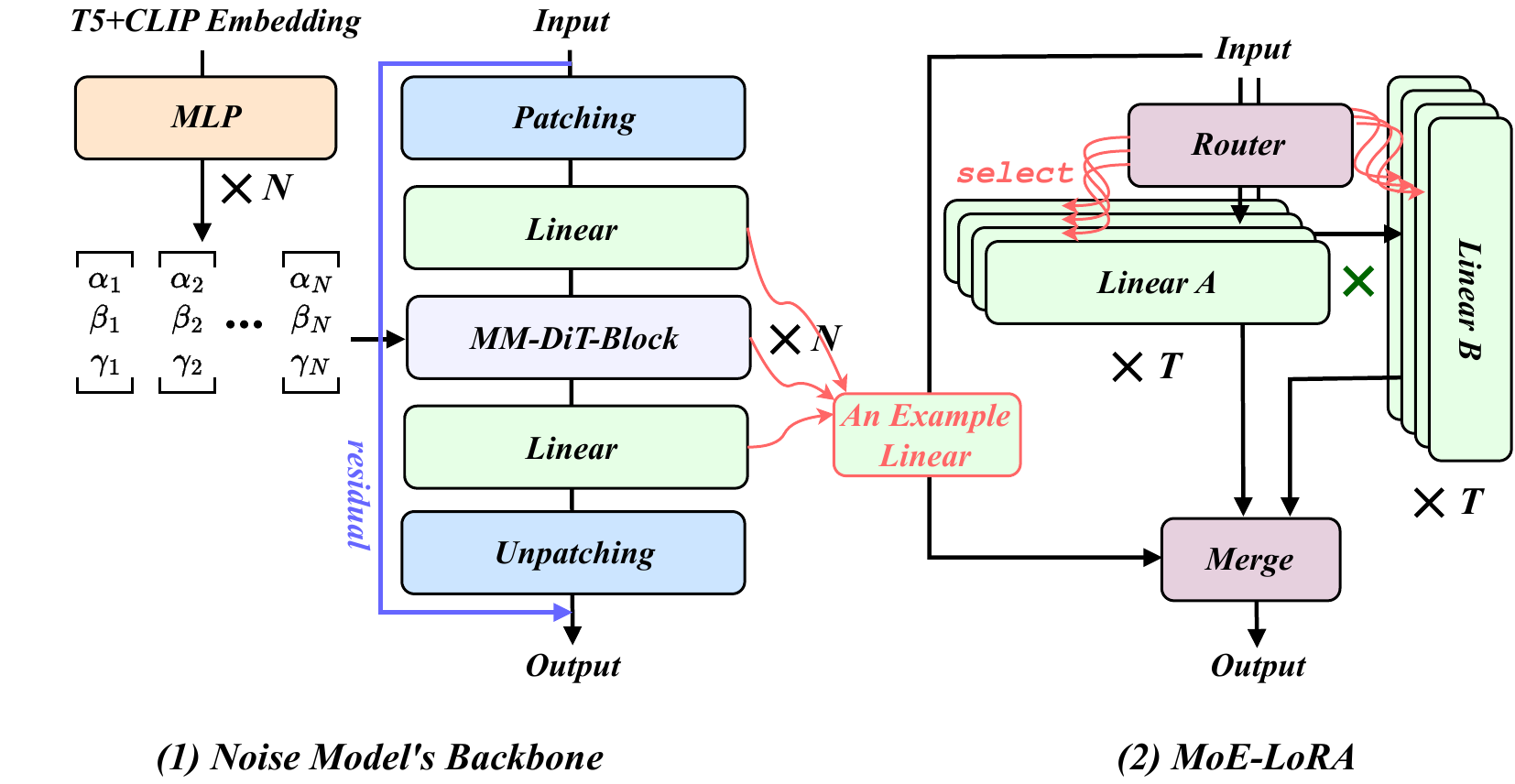}
    \vspace{-18pt}\caption{Framework of noise model's backbone and MoE-LoRA. We employ MM-DiT-Block to construct noise model for SD3.5 and SDXL, and we utilize MoE-LoRA to reflect easy-to-learn content across different timesteps. For LlamaGen, we replace the backbone with its native Llama block~\citep{llama} w/o MoE-LoRA.}
    \label{fig:moe_lora}
    \vspace{-15pt}
\end{figure}

\vspace{-7pt}\paragraph{Stage-\Rmnum{2} \textit{Reflect} $\left(\imineq{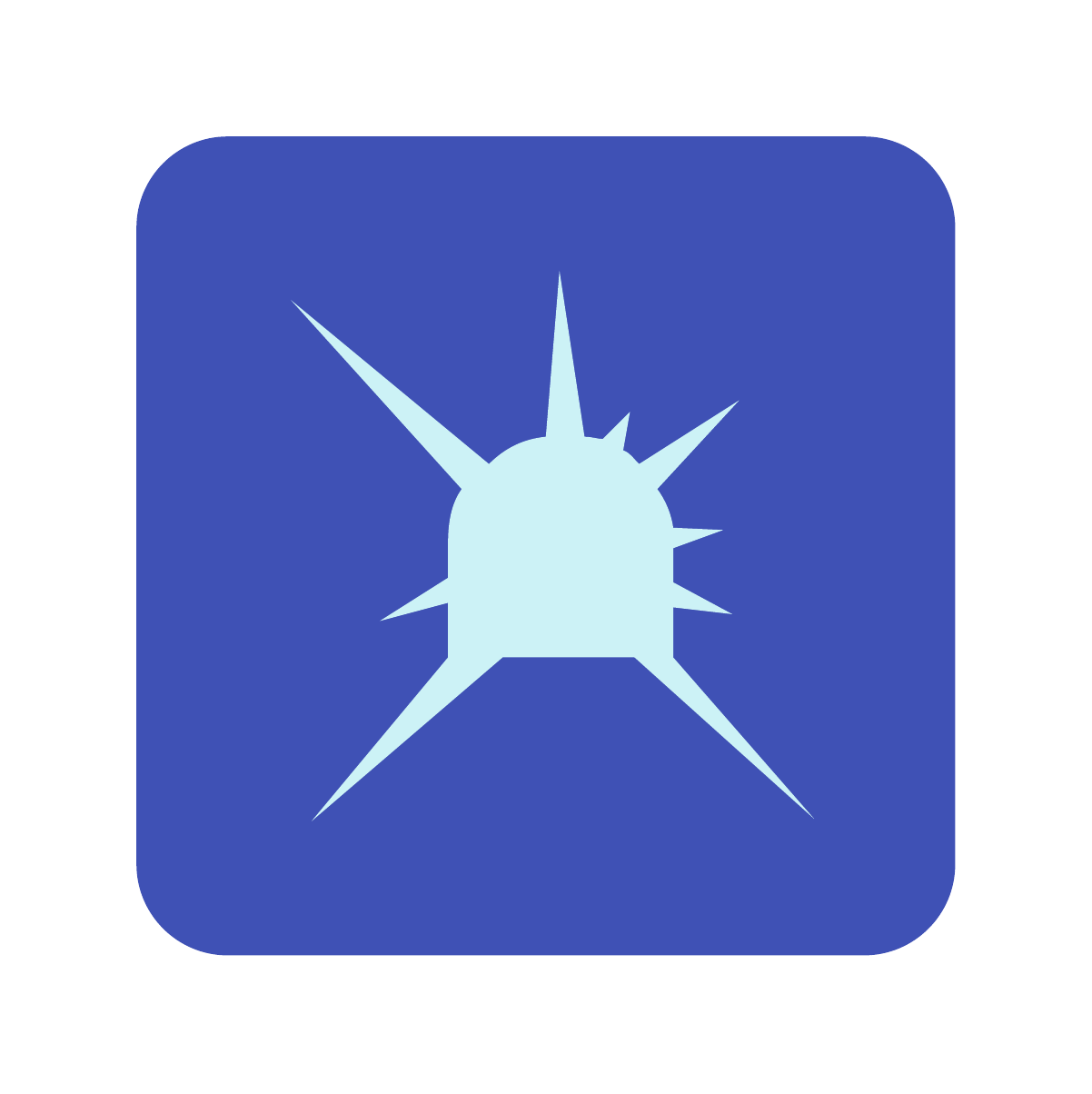}{2.4}\right)$.} This phase is the most critical step, as the weak model trained here ultimately determines the success of the subsequent \textit{Refine} stage. We require the weak model to perform comparably to the strong model in reflecting the easy-to-learn aspects of the image, while allowing a noticeable gap to remain between the two when it comes to capturing the difficult-to-learn components. To achieve this, we select noise model with a lightweight architecture and relatively limited representational capacity. As illustrated in Fig.~\ref{fig:moe_lora} (1), we utilize the MM-DiT-Block~\citep{FLUX} to construct the noise model. We lower the amount of layers in the architecture to reduce GPU delay. Additionally, we use the weight selection technique~\citep{weight_selection} to initialize the weak model, which speeds up the training process. However, noise model struggles to generalize effectively across different sampling steps. Thus, we enhance the existing weak model component by integrating low-rank adaptation (LoRA)~\citep{hu2022lora} for each sampling step, thereby constructing MoE-LoRA~\citep{moelora} (see Fig.~\ref{fig:moe_lora} (2)). This technique significantly improves the weak model's generalization ability. Finally, the reflect loss can be denoted as
\begin{equation}
\fontsize{7pt}{11pt}\selectfont
    \begin{split}
        & \mathcal{L}_\text{reflect}=\alpha\mathbb{E}_{(t,i)}{\cos\left(\frac{T-t}{T}\right)}\left\Vert\epsilon_\theta^\textrm{weak}(\epsilon_{\textrm{uncond}}^{(t,i)}, \mathcal{C}^{(t,i)}) - \epsilon_{\textrm{cond}}^{(t,i)}\right\Vert_2^2,\\
    \end{split}
\end{equation}where {\footnotesize \( \cos\left(\frac{\alpha(T-t)}{T}\right) \)} represents dynamic weights, designed to encourage the noise model to focus more on the critical initial sampling stages~\citep{shao2025ivmixed,shao2024the}, and the hyperparameter \( \alpha \) is set to a default value of 4. 
\vspace{-3pt}
\paragraph{Stage-\Rmnum{3} \textit{Refine} $\left(\imineq{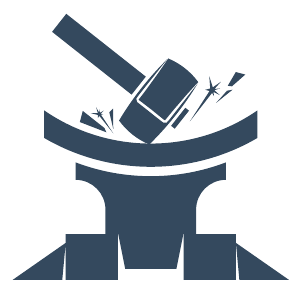}{2.4}\right)$.} After obtaining the well-trained weak model $\epsilon_\theta^\textrm{weak}$, we can leverage it for both fast and enhanced inference (i.e., fast and slow mode in Fig.~\ref{fig:total_framework_core2}). Inspired by~\citep{bad_version_diffusion}, we adopt W2S guidance in the slow mode. However,~\citep{bad_version_diffusion} only replaces the unconditional output with the weak output, we further extend this by substituting the noise model output with the weak output and the CFG output with the strong output. Consequently, the slow mode and the fast mode are ($\omega_\textrm{w2s}$ is the W2S guidance scale)
\begin{equation}
    \begin{split}
        \epsilon^t_\textrm{slow-mode} =& \epsilon^t_\textrm{weak} + \omega_\textrm{w2s}(\epsilon^t_\textrm{strong} - \epsilon^t_\textrm{weak}),\\
        \epsilon^t_\textrm{fast-mode}\  =& \epsilon^t_\textrm{weak},\\
    \end{split}
    \label{eq:w2s_guidance}
\end{equation}where $\epsilon^t_\textrm{weak}$ denotes the output of $\epsilon_\theta^\textrm{weak}$ at timestep $t$. In practical implementation, we can use \(\epsilon^t_\textrm{slow-mode}\) to enhance inference or \(\epsilon^t_\textrm{fast-mode}\) to accelerate it. Specifically, the slow mode is applied in the early stages of sampling, while the fast mode is employed during the later stages. As shown in Fig.~\ref{fig:first_presentation}, by balancing the number of invocations of these two modes, we achieve acceleration of 1.31s and 2.89s on SDXL and SD3.5, respectively, under the same performance. Meanwhile, under the same GPU latency, we observe improvements of 0.39 and 0.47 in HPS v2~\citep{HPSV2} on SDXL and SD3.5, respectively. We will go on to describe W2S guidance from both a theoretical and empirical standpoint in order to comprehend how it improves inference.

\noindent{\bf Theoretical Analysis.} Under Eq.~\ref{eq:w2s_guidance} we give this definition:
\begin{definition}
\vspace{-4pt}
\label{sec:def1}
(the formal definition in Def.~\ref{apd:def1}) Let a Gaussian mixture model (GMM) \( g(x) = \sum_{i=1}^M w_i \mathcal{N}(\mu_i, \Sigma_i) \) be the ground truth score function in DM with \( \sum_{i=1}^M w_i = 1 \) at the timestep $t$. We partition the components of the score function estimation \( \tilde{g}(x) = \sum_{i=1}^M w_i \mathcal{N}(\tilde{\mu}_i, \tilde{\Sigma}_i) \) into ``easy-to-learn'' and ``difficult-to-learn'' subsets based on their approximation errors ($M_1$ is a hyperparameter):

\noindent{\bf\textit{Easy-to-Learn Components (\( i \in \{1,\ldots,M_1\} \)).}} The approximation error for these components satisfies: $\sum_{i=1}^{M_1} w_i \|\mu_i - \tilde{\mu}_i\|^2 \leq \eta_{\text{easy}}$.

\noindent{\bf\textit{Difficult-to-Learn Components (\( i \in \{M_1+1,\ldots,M\} \)).}} These components exhibit larger approximation errors: $\eta_{\text{easy}} < \sum_{i=M_1+1}^M w_i \|\mu_i - \tilde{\mu}_i\|^2 \leq \eta_{\text{difficult}}$.

\end{definition}

\noindent{Through Definition~\ref{sec:def1} (using DM as an example), \(\epsilon^t_\textrm{weak}\) and \(\epsilon^t_\textrm{strong}\) can be modeled as the estimator \( \tilde{g}(x) = \sum_{i=1}^M w_i \mathcal{N}(\tilde{\mu}_i, \tilde{\Sigma}_i) \). Now we can establish this theorem:}

\begin{theorem}
\label{the:why_w2s_guidance_work}
(the proof in Appendix~\ref{apd:do_w2s_work}) Assume there are two macro-level DMs, denoted as \( \epsilon_\theta^\textrm{weak} \) and \( \epsilon_\theta^\textrm{strong} \). According to Definition~\ref{sec:def1} (the symbols in Eq.~\ref{eq:condition_main_paper} have the same meaning as in Definition~\ref{sec:def1} when the right superscripts ``weak'' and ``strong'' are removed), these models handle ``easy-to-learn'' and ``difficult-to-learn'' content, where it is assumed that the contents overlap. The key distinction (i.e., constraints) is as follows:
\begin{equation}
\footnotesize
    \begin{split}
        & |\eta_\textrm{easy}^\textrm{weak} - \eta_\textrm{easy}^\textrm{strong}| < |\eta_\textrm{difficult}^\textrm{weak} - \eta_\textrm{difficult}^\textrm{strong}|, \eta_\textrm{difficult}^\textrm{strong} < \eta_\textrm{difficult}^\textrm{weak}, \\
        &\eta_\textrm{easy}^\textrm{strong} < \eta_\textrm{easy}^\textrm{weak},(\mu_i - \mu_i^\textrm{strong})(\mu_i^\textrm{strong} - \mu_i^\textrm{weak}) > 0.\\
    \end{split}
    \label{eq:condition_main_paper}
\end{equation} There exists $\omega_\textrm{w2s} > 1$ (i.e., the W2S guidance scale) such that the mean square error between W2S guidance $\epsilon_\theta^\textrm{weak}(x_t,t) + \omega_\textrm{w2s}\left[\epsilon_\theta^\textrm{strong}(x_t,t)- \epsilon_\theta^\textrm{weak}(x_t,t)\right]$ and the ground truth distribution $\sum_{i=1}^M w_i \mathcal{N}(\mu_i, \Sigma_i)$ is smaller than either $\Vert\sum_{i=1}^M w_i \mathcal{N}(\mu_i, \Sigma_i)-\epsilon_\theta^\textrm{weak}(x_t,t)\Vert_2^2$ or $\Vert \sum_{i=1}^M w_i \mathcal{N}(\mu_i, \Sigma_i)-\epsilon_\theta^\textrm{strong}(x_t,t)\Vert_2^2$ alone.
\vspace{-6pt}
\end{theorem}In Theorem~\ref{the:why_w2s_guidance_work}, \(\epsilon^t_\textrm{weak}\) and \(\epsilon^t_\textrm{strong}\) correspond to the outputs of \(\epsilon_\theta^\textrm{weak} \) and \(\epsilon_\theta^\textrm{strong} \). Thus, there is always an optimal \(\omega_\textrm{w2s}>1\) that can improve the inference performance.\begin{figure}
    \centering
    \includegraphics[width=1.0\linewidth]{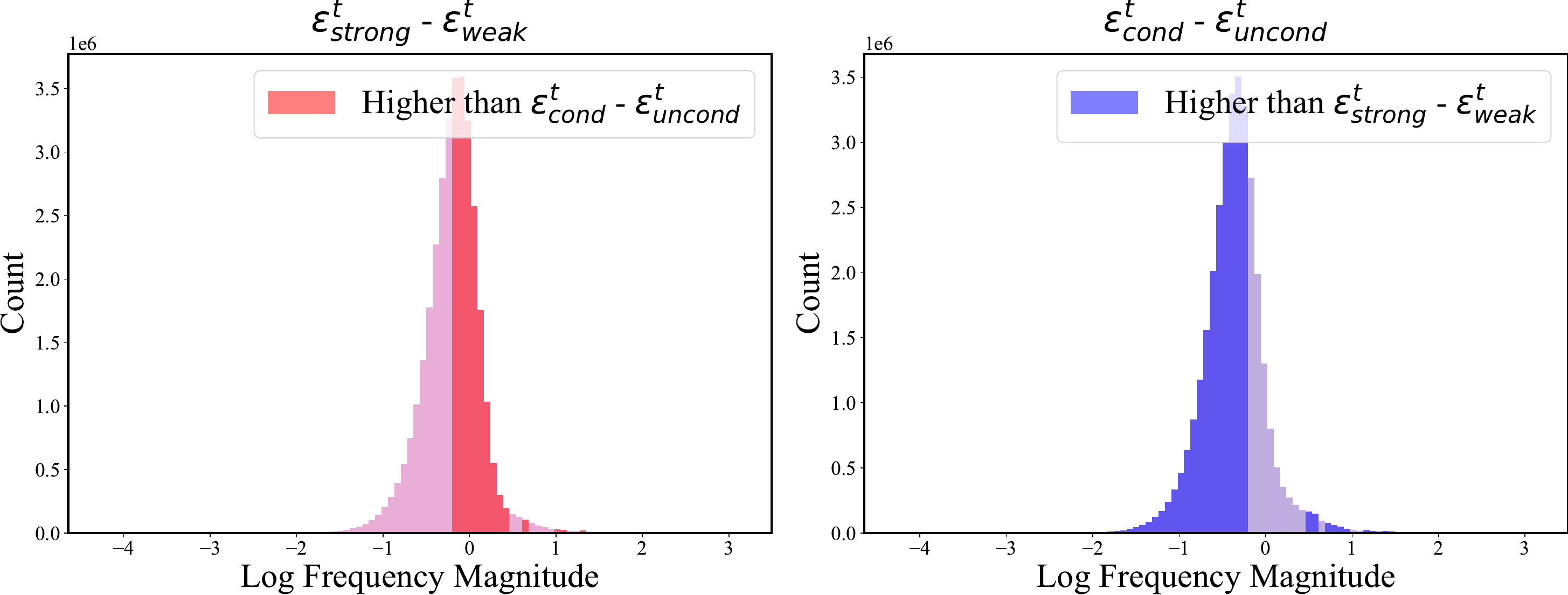}
    \vspace{-10pt}
    \caption{Visualization of the frequency histogram with the W2S guidance $\epsilon^t_\textrm{strong}\!-\!\epsilon^t_\textrm{weak}$ and CFG $\epsilon^t_\textrm{cond}\!-\!\epsilon^t_\textrm{uncond}$ in SDXL. Note that W2S guidance in \texttt{CoRe$^2$} incorporates more high-frequency information compared to standard CFG. This effectively mitigates the base model's limitations in generating fine-grained yet challenging-to-learn content during the pre-training phase.}
    \label{fig:frequency_core2}
    \vspace{-13pt}
\end{figure}\begin{figure}
    \centering
    \includegraphics[width=1.0\linewidth]{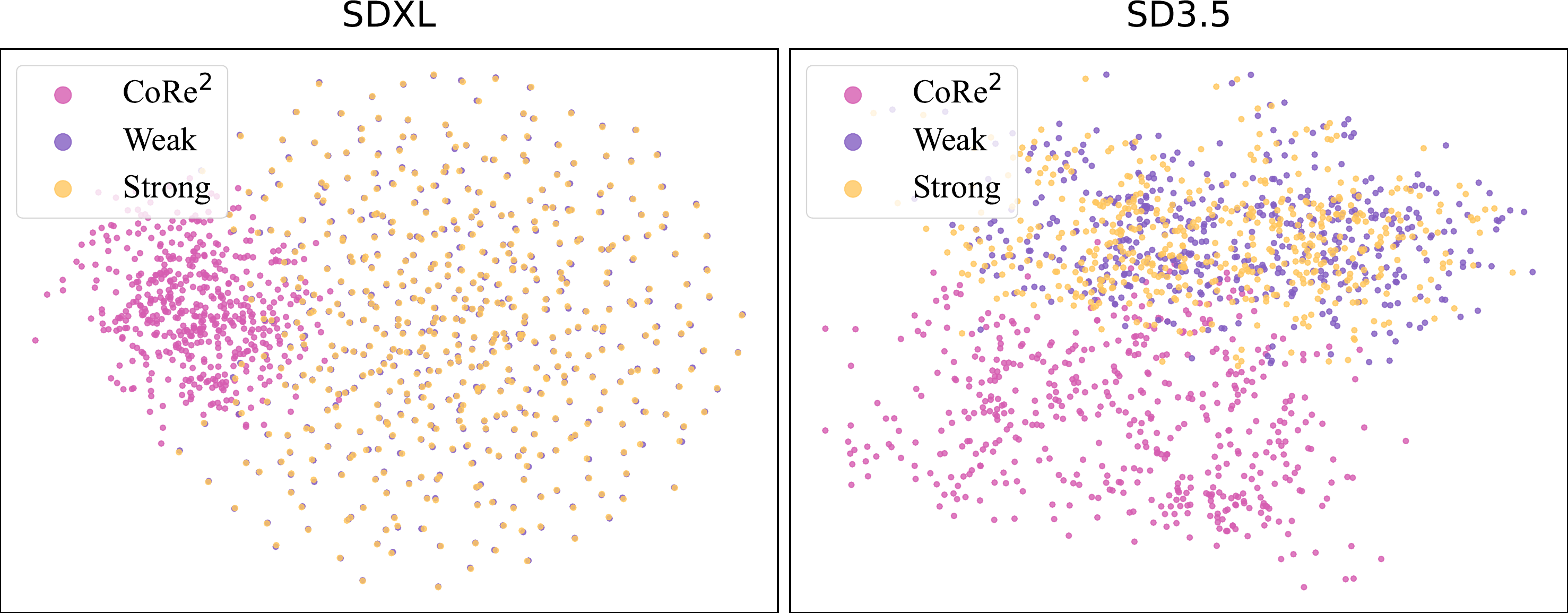}
    \vspace{-10pt}
    \caption{Visualization of the model output obtained by \texttt{CoRe$^2$}, compared with those from the strong model (i.e., standard CFG) and the weak model (i.e., the noise model). We begin by generating 500 initial Gaussian noise samples and their corresponding model outputs for the text prompt ``olympic swimming pool''. These are then projected into a lower-dimensional space using T-SNE~\citep{van2008visualizing} to produce the clear visualization.}
    \label{fig:t_sne_core2}
    \vspace{-19pt}
\end{figure} 

\noindent{\bf Empirical Understanding.} We use Fig.~\ref{fig:frequency_core2} to illustrate the differences in guidance direction between W2S guidance and the classical CFG. It can be observed that, compared to the classical CFG, W2S guidance encapsulates more high-frequency information, which contributes to refining details and textures within the image, thereby significantly enhancing image quality. Coupled with Definition~\ref{sec:def1}, these high-frequency components can be identified as the so-called "difficult-to-learn" elements. Furthermore, as shown in Fig.~\ref{fig:empirical_vis_core2}, the distribution of \(\epsilon_\textrm{slow-mode}^t\) (i.e., \texttt{CoRe}$^2$ in Fig.~\ref{fig:empirical_vis_core2}) is noticeably different from those of \(\epsilon_\textrm{weak}^t\) (i.e., Weak in Fig.~\ref{fig:empirical_vis_core2}) and \(\epsilon_\textrm{strong}^t\) (i.e., Strong in Fig.~\ref{fig:empirical_vis_core2}). This observation further indicates that W2S guidance indeed introduces significant corrections to the synthesized image, which, in our experiments, manifest as notable performance improvements.

\input{tables/zigzag_comparison}
\vspace{-6pt}
\paragraph{\texttt{Z-CoRe}$^2$.} Note that the form of W2S guidance is not limited to Eq.~\ref{eq:w2s_guidance}. By integrating it with the resampling algorithm Z-Sampling~\citep{zigzag}, we derive a novel algorithm, \texttt{Z-CoRe}$^2$, which achieves superior performance compared to Z-Sampling while incurring lower inference costs. The algorithmic logic is outlined as\begin{equation}
    \begin{split}
        x_{t-1} &= \textrm{DDIM}(x_t, \textrm{slow-mode}),\\
        x_t &= \textrm{DDIM-Inversion}(x_{t-1}, \textrm{fast-mode}),\\
        x_{t-1} &= \textrm{DDIM}(x_{t}, \emptyset),\\
    \end{split}
\end{equation}
where DDIM$(\cdot,\cdot)$ and DDIM-Inversion$(\cdot,\cdot)$ stand for the DDIM~\citep{ddim} and DDIM Inversion~\citep{ddiminversion} sampling technique, respectively, and fast-mode, slow-mode and $\emptyset$ represent (reverse) fast mode in Fig.~\ref{fig:total_framework_core2}, (forward) slow mode in Fig.~\ref{fig:total_framework_core2} and (forward) DDIM with the standard CFG, respectively.

%% file: tables/zigzag_comparison.tex
\begin{table}[!t]
\centering
\vskip -0.01in
\small
\caption{Comparison between \texttt{Z-CoRe}$^2$ and Z-Sampling. \texttt{Z-CoRe}$^2$ demonstrates both faster and superior performance compared to Z-Sampling.}  
\scalebox{0.7}{
\setlength{\tabcolsep}{12pt} 
\renewcommand{\arraystretch}{0.85} 
\begin{tabular}{lccc}
\Xhline{3\arrayrulewidth}\\[-2ex]
{\textbf{Method}} & {\textbf{PickScore (↑)}} & {\textbf{AES (↑)}} & {\textbf{GPU Latency (↓)}} \\
\Xhline{3\arrayrulewidth}
        \\[-2ex]
        \multicolumn{4}{l}{\large\textit{\textbf{SDXL}}}\\\Xhline{3\arrayrulewidth}
Pick-of-Pic & & & \\
\quad Z-Sampling & 21.94 & 5.98 & 25.7653 \\
\quad \texttt{Z-CoRe}$^2$ (Ours) & \CC{21.95} & \CC{6.06} & \CC{23.7565} \\[0.5ex]
DrawBench & & & \\
\quad Z-Sampling & 23.10 & 5.69 & 25.7653 \\
\quad \texttt{Z-CoRe}$^2$ (Ours) & \CC{23.20} & \CC{5.71} & \CC{23.7565} \\

\multicolumn{4}{l}{\large\textit{\textbf{SD3.5}}}\\\Xhline{3\arrayrulewidth}
Pick-of-Pic & & & \\
\quad Z-Sampling & 21.81 & 5.81 & 63.1081 \\
\quad \texttt{Z-CoRe}$^2$ (Ours) & \CC{22.11} & \CC{5.97} & \CC{57.4656} \\[0.5ex]
DrawBench & & & \\
\quad Z-Sampling & 22.67 & 5.48 & 63.1081 \\
\quad \texttt{Z-CoRe}$^2$ (Ours) & \CC{22.83} & \CC{5.57} & \CC{57.4656} \\
\Xhline{3\arrayrulewidth}\\[-2ex]
\end{tabular}
}
\label{tab:comparison_z_Sampling}
\vskip -0.07in
\end{table}

%% file: sec/4_experiment.tex
\section{Experiments}
\label{sec:experiment}
\input{tables/main_comparison}
\input{tables/main_geneval}
\input{tables/main_t2i_compbench}

\begin{figure*}[t]
    \vspace{-5pt}
    \centering
    \includegraphics[width=1.0\linewidth]{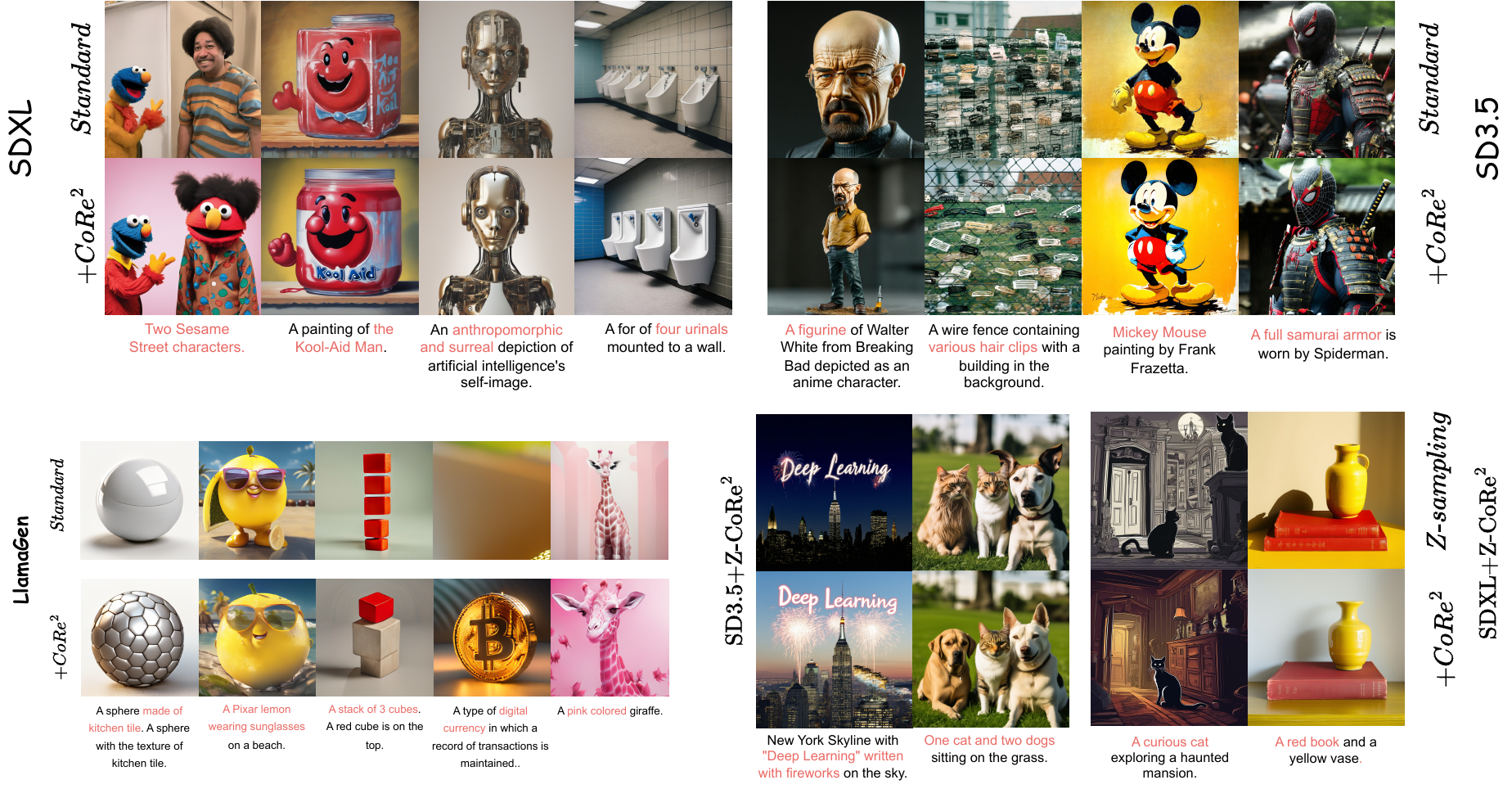}
    \vspace{-29pt}\caption{Visualization of the standard sampling, \texttt{CoRe}$^2$ and \texttt{Z-CoRe}$^2$ on SD3.5, SDXL and LlamaGen. Compared to standard sampling, \texttt{CoRe}$^2$ and \texttt{Z-CoRe}$^2$ excel in generating images with more intricate high-frequency details, while also improving semantic consistency. In Appendix~\ref{apd:additional_visualization}, we present additional visualizations of \texttt{CoRe}$^2$ across four distinct styles: anime, concept art, painting, and photography.}
    \label{fig:main_visualization}
    \vspace{-14pt}
\end{figure*}\begin{figure}[t]
    \centering
    \vspace{-8pt}
    \includegraphics[width=0.8\linewidth]{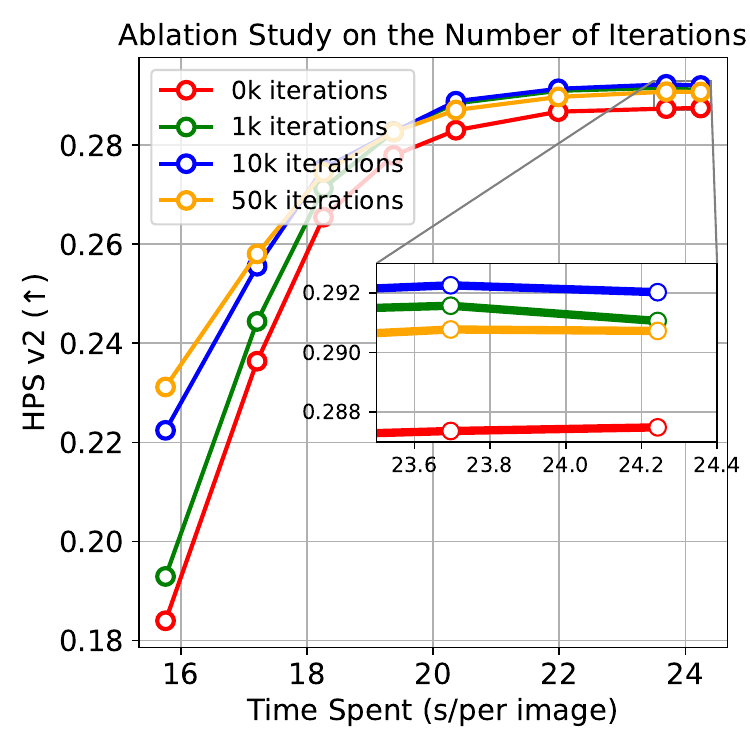}
    \vspace{-15pt}
    \caption{Ablation study of iteration numbers during the \textit{Reflect} phase using SD3.5. We find that the 10k iterations version contribute the best performance to the full-capacity \texttt{CoRe}$^2$.}
    \label{fig:ablation_study_sd35_1k_10k_50k}
    \vspace{-21pt}
\end{figure}\noindent{\bf Benchmark, Evaluation Metric and Dataset.} In this paper, we utilize several popular benchmarks, including Pick-of-Pic~\citep{pick_of_pic}, DrawBench~\citep{drawbench}, HPD v2~\citep{HPSV2}, GenEval~\citep{geneval}, and T2I-Compbench~\citep{t2i_compbench}. For Pick-of-Pic, DrawBench, and HPD v2, we evaluate performance using four distinct metrics: PickScore~\citep{pick_of_pic}, AES~\citep{AES}, HPS v2~\citep{HPSV2}, and ImageReward~\citep{Imagereward}. For GenEval and T2I-Compbench, we follow the official evaluation protocols, which assess various dimensions such as object color, color binding, shape binding, and texture binding. Regarding the synthesized datasets obtained during the \textit{Collect} phase, we gather 200k, 100k, and 100k samples on SD3.5~\citep{SD35}, SDXL~\citep{SDXL}, and LlamaGen~\citep{llama}, respectively. Details about benchmarks, evaluation metrics and datasets can be found in Appendix~\ref{apd:benchmark_and_dataset}. 

\noindent{\bf Our Implementation.} Three essential elements make up our implementation: \textit{Collect}, \textit{Reflect}, and \textit{Refine}. For the \textit{Collect} phase, we employ the default CFG scale, and all prompts are refined using InternVL2-26b~\citep{internvl2} to ensure the dataset's quality. In the \textit{Reflect} phase, we utilize the AdamW optimizer. Unless otherwise specified, the learning rate is set to 2e-6, with a batch size of 64 and 10k iterations. In the \textit{Refine} phase, we utilize the default slow mode for the full sampling path in all experimental results presented in the tables, unless specified otherwise. Additionally, the ablation results illustrated in the figures are obtained by balancing the utilization frequencies of the fast mode and the slow mode. Please check Appendix~\ref{apd:detail_implementation} for a more comprehensive implementation.

\subsection{Main Results}
\label{sec:main_result}
Our results on SDXL, SD3.5 and LlamaGen are presented below, and FLUX's results can be found in Appendix~\ref{apd:flux_experiment}.

\noindent{\bf SDXL Comparison.} SDXL is a seminal and widely recognized generative model in the I2V research domain. To validate that \texttt{CoRe}$^2$ can enhance both the quality and realism of synthesized images, and demonstrate its superiority over other inference-enhanced algorithms, we present our experimental results in Tables~\ref{tab:comparison_main},~\ref{tab:comparison_geneval} and~\ref{tab:comparison_t2i_compbench}. Table~\ref{tab:comparison_main} illustrates a performance comparison between \texttt{CoRe}$^2$ and other SOTA algorithms on the Pick-of-Pic, DrawBench, and HPD v2 benchmarks. The evaluated SOTA methods include CFG++~\citep{chung2024cfg++}, Guidance Interval~\citep{guidance_interval}, and PAG~\citep{pag_diffusion}. As shown, \texttt{CoRe}$^2$ consistently surpasses the competing methods across all metrics, with the sole exception of AES. Notably, this superior performance is achieved with only an additional \~0.21s delay compared to the standard sampling. Furthermore, \texttt{CoRe}$^2$ demonstrates impressive performance on both GenEval and T2I-Compbench, surpassing the standard sampling by margins of 1.62\% and 1.52\% on the overall metrics, respectively.

\begin{figure*}
    \centering
    \includegraphics[width=1.0\linewidth]{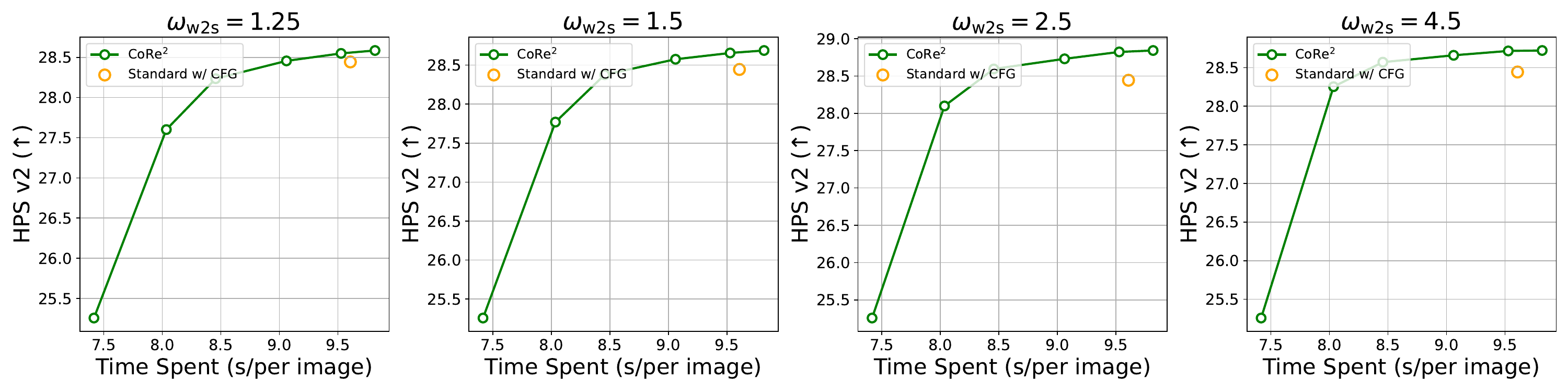}
    \vspace{-23pt}
    \caption{Ablation study of \( \omega_\textrm{w2s} \) on SDXL. We observe that \texttt{CoRe}$^2$ achieves optimal performance when \( \omega_\textrm{w2s} = 2.5 \).}
    \label{fig:ablation_study_sdxl}
        \vspace{-13pt}
\end{figure*}\begin{figure*}
    \centering
    \includegraphics[width=1.0\linewidth]{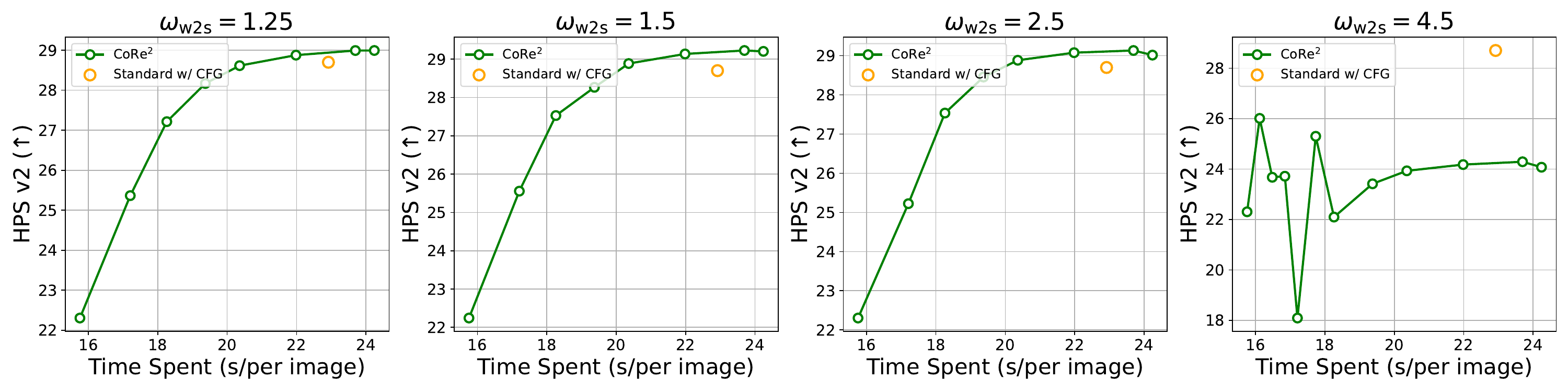}
    \vspace{-23pt}
    \caption{Ablation study of \( \omega_\textrm{w2s} \) on SD3.5. We observe that \texttt{CoRe}$^2$ achieves optimal performance when \( \omega_\textrm{w2s} = 1.5 \).}
    \label{fig:ablation_study_sd35}
        \vspace{-18pt}
\end{figure*}

\noindent{\bf SD3.5 Comparison.} SD3.5 is a large-scale I2V DM that use flow matching as its noise schedule. In our experiments, we observe that many algorithms that perform effectively on SDXL fail to deliver similar results on SD3.5. In detail, the comparative results are presented in Tables~\ref{tab:comparison_main},~\ref{tab:comparison_geneval} and~\ref{tab:comparison_t2i_compbench}. Beyond the comparison methods considered on SDXL, we additionally include Z-Sampling~\citep{zigzag} for evaluation. It can be observed that for Pick-of-Pic, DrawBench, and HPD v2, \texttt{CoRe}$^2$ significantly outperforms the compared methods across all metrics except AES. This superior performance comes with only a 5.7\% increase in computational overhead compared to the standard sampling. Similar to its performance on SDXL, \texttt{CoRe}$^2$ demonstrates significant improvements on both GenEval and T2I-Compbench, achieving an average performance gain of 1.52\% and 0.54\%, respectively, compared to the standard sampling.

\noindent{\bf LlamaGen Comparison.} LlamaGen is the first prominent visual ARM capable of scaling to a resolution of 512$\times$512. Demonstrating the effectiveness of \texttt{CoRe}$^2$ on LlamaGen further highlights the adaptability of our approach. However, due to the decoder-only Transformer architecture employed in our noise model, resulting in slightly higher latency compared to SDXL and SD3.5. As shown in Fig.~\ref{fig:first_presentation}, under identical GPU latency conditions, \texttt{CoRe}$^2$ does not outperform the standard sampling. Nevertheless, the performance scaling laws of \texttt{CoRe}$^2$ remain valid on LlamaGen. As illustrated in Table~\ref{tab:comparison_main}, although PickScore on Pick-of-Pic is marginally lower than that of the standard sampling, \texttt{CoRe}$^2$ achieves substantial improvements across all other metrics and benchmarks. These findings provide strong evidence of the robust generalization capabilities of \texttt{CoRe}$^2$.

\noindent{\bf \texttt{Z-CoRe}$^2$ vs. Z-Sampling.} As mentioned in Sec.~\ref{sec:method}, \texttt{CoRe}$^2$ can be extended with Z-Sampling to achieve W2S guidance, resulting in \texttt{Z-CoRe}$^2$. We present the comparison results of \texttt{Z-CoRe}$^2$ and Z-Sampling on SDXL (50 steps) and SD3.5 (28 steps) in Table~\ref{tab:comparison_z_Sampling}. The results clearly demonstrate that \texttt{Z-CoRe}$^2$ not only outperforms Z-Sampling in terms of effectiveness but also achieves higher efficiency.

\subsection{Ablation Study}
\label{sec:ablation_study}

\noindent{\bf Number of Iterations during the \textit{Reflect} Phase.} From Fig.~\ref{fig:ablation_study_sd35_1k_10k_50k}, we observe that as the number of iterations in the \textit{Reflect} stage increases from 1k to 10k and then to 50k, the performance of the full sampling path using fast mode (i.e., the leftmost three points) progressively improves, indicating that the weak model's capability is gradually enhanced. It is important to note that 0k represents the direct use of the conditional output as the result for the weak output. This approach is evidently less effective compared to its counterparts, such as 1k, 10k, and 50k. Nevertheless, the best performance is achieved at the 10k iterations version (i.e., the blue line), suggesting that the weak model should ideally remain moderately ``weak'', striking a balance rather than continually pushing its performance ceiling.


\noindent{\bf W2S Guidance Scale $\omega_\textrm{w2s}$.} In Fig.~\ref{fig:ablation_study_sdxl} and~\ref{fig:ablation_study_sd35}, we provide the ablation study results for $\omega_\textrm{w2s}$ on SDXL and SD3.5, respectively, and the corresponding analysis for LlamaGen is detailed in Appendix~\ref{apd:w2s_llamagen_ablation_study}. As $\omega_\textrm{w2s}$ increases, the performance of \texttt{Z-CoRe}$^2$ follows a trend of initial improvement, peaking before subsequently declining. The empirical optimal values are found to be 1.5 for SD3.5 and 2.5 for SDXL, indicating the importance of carefully tuning $\omega_\textrm{w2s}$ for different models. Furthermore, for LlamaGen, we determine that the optimal performance is achieved when $\omega_\textrm{w2s}$=1.5.

\noindent{\bf Visualization.} In Fig.~\ref{fig:main_visualization} and Appendix~\ref{apd:additional_visualization}, we showcase images synthesized by \texttt{CoRe}$^2$, \texttt{Z-CoRe}$^2$, and the standard sampling. It is evident that both \texttt{CoRe}$^2$ and \texttt{Z-CoRe}$^2$ significantly enhance detail textures, realism, and semantic faithfulness compared to the standard sampling and Z-Sampling. Furthermore, on LlamaGen, \texttt{CoRe}$^2$ demonstrates the capability to address out-of-domain challenges for prompts such as ``A type of digital...is maintained'', showcasing its robustness in diverse scenarios.

%% file: tables/main_comparison.tex
\begin{table*}[!t]
\centering
\vspace{-8pt}
\small
\caption{Comparison between \texttt{CoRe}$^2$ and other SOTA methods on Pick-of-Pic, DrawBench and HPD v2. The full-capacity \texttt{CoRe}$^2$ demonstrates outstanding performance and it incurs only a negligible additional delay compare with the standard sampling.}
\vspace{-8pt}
\setlength{\tabcolsep}{20pt} 
\renewcommand{\arraystretch}{0.85} 
\scalebox{0.8}{
\begin{tabular}{lcccccr}
\Xhline{3\arrayrulewidth}\\[-2ex]
{\textbf{Method}} & {\textbf{PickScore (↑)}} & {\textbf{HPSv2 (↑)}} & {\textbf{AES (↑)}} & {\textbf{ImageReward (↑)}} & {\textbf{GPU Latency (↓)}}\\
\Xhline{3\arrayrulewidth}
        \\[-2ex]
        \multicolumn{6}{l}{\large\textit{\textbf{SDXL}}}\\\Xhline{3\arrayrulewidth}
Pick-a-Pic & & & & & \\
\quad Standard & 21.7403 & 27.4275 & 6.0342 & 0.5255 & 9.6066 \\
\quad CFG++ & 21.0357 & 27.5571 & 5.9449 & 0.4215 & 9.6066 \\
\quad Guidance Interval & 20.1426 & 22.3632 & 5.9763 & -0.4111 & 9.5814 \\
\quad PAG & 21.0409 & 27.3398 & \CC{6.0608} & 0.2603 & 12.5134 \\
\quad \texttt{CoRe}$^2$ (Ours) & \CC{21.7967} & \CC{27.9911} & {5.9057} & \CC{0.7042} & {9.8169} \\\hline
Drawbench & & & & & \\
\quad Standard & 22.2300 & 28.5300 & 5.5880 & 0.5441 & 9.6066 \\
\quad CFG++ & 22.2400 & 28.8100 & 5.5720 & 0.6333 & 9.6066 \\
\quad Guidance Interval & 21.4100 & 26.7600 & 5.4670 & 0.1501 & 9.5814 \\
\quad PAG & 22.1200 & 28.4800 & \CC{5.7030} & 0.4349 & 12.5134 \\
\quad \texttt{CoRe}$^2$ (Ours) &  \CC{22.3433} &  \CC{28.9763} &  {5.5475} &  \CC{0.7233} &  {9.8169} \\\hline
HPD v2 & & & & & \\
\quad Standard & 22.4600 & 28.4400 & \CC{6.1000} & 0.8889 & 9.6066 \\
\quad \texttt{CoRe}$^2$ (Ours) & \CC{22.4900} & \CC{28.8400} & {5.9800} & \CC{0.9600} & 9.8169 \\

\multicolumn{6}{l}{\large\textit{\textbf{SD3.5}}}\\\Xhline{3\arrayrulewidth}
Pick-a-Pic & & & & & \\
\quad Standard & 22.0661 & 27.9503 & 5.8669 & 0.8411 & 22.9196 \\
\quad CFG++ & 21.9211 & 27.6814 & 5.8211 & 0.7850 & 22.9196 \\
\quad Guidance Interval & 22.0497 & 27.7957 & 5.8738 & 0.7989 & 21.8679 \\
\quad PAG & 19.5623 & 23.9618 & 4.8977 & -1.0452 & 31.7245 \\
\quad Z-Sampling & 21.3563 & 26.8228 & 5.8397 & 0.6510 & 26.8410 \\
\quad \texttt{CoRe}$^2$ (Ours) & \CC{22.0876} & \CC{28.1279} & \CC{5.8948} & \CC{0.8924} & 24.2429 \\\hline
Drawbench & & & & & \\
\quad Standard & 22.7461 & 29.0764 & 5.4901 & 0.8802 & 22.9196 \\
\quad CFG++ & 22.6327 & 28.7629 & 5.4205 & 0.8082 & 22.9196 \\
\quad Guidance Interval & 22.7739 & 29.0078 & \CC{5.5050} & 0.8768 & 21.8679 \\
\quad PAG & 20.5133 & 25.0731 & 4.8975 & -1.0572 & 31.7245 \\
\quad Z-Sampling & 22.4124 & 28.5431 & 5.4734 & 0.7694 & 26.8410 \\
\quad \texttt{CoRe}$^2$ (Ours) & \CC{22.7800} & \CC{29.2160} & {5.4926} & \CC{0.9168} & 24.2429 \\\hline
HPD v2& & & & & \\
\quad Standard & 22.6012 & 28.8000 & 5.7618 & 0.9860 & 22.9196 \\
\quad \texttt{CoRe}$^2$ (Ours) & \CC{22.6445} & \CC{29.2030} & \CC{5.7623} & \CC{1.0813} & 24.2429 \\

\multicolumn{6}{l}{\large\textit{\textbf{LlamaGen}}}\\\Xhline{3\arrayrulewidth}
Pick-a-Pic & & & & & \\
\quad Standard & \CC{19.9200} & 25.1900 & 5.7600 & -0.5500 & 44.4923 \\
\quad \texttt{CoRe}$^2$ (Ours) & 19.8600 & \CC{25.2200} & \CC{5.8600} & \CC{-0.4700} & 48.6858 \\\hline
Drawbench & & & & & \\
\quad Standard & 20.6200 & 26.3100 & 5.6900 & -0.5000 & 44.4923 \\
\quad \texttt{CoRe}$^2$ (Ours) & \CC{20.7800} & \CC{26.5900} & \CC{5.8100} & \CC{-0.3700} & 48.6858 \\\hline
HPD v2& & & & & \\
\quad Standard & 20.7300 & 26.6700 & 6.0600 & 0.0400 & 44.4923 \\
\quad \texttt{CoRe}$^2$ (Ours) & \CC{20.7700} & \CC{26.8800} & \CC{6.1400} & \CC{0.1200} & 48.6858 \\
\Xhline{3\arrayrulewidth}\\[-2ex]
\end{tabular}
}
\label{tab:comparison_main}
\vspace{-10pt}
\end{table*}

%% file: tables/main_geneval.tex
\begin{table*}[!t]
\centering
\vskip -0.01in
\small
\caption{Comparison between \texttt{CoRe}$^2$ and the standard sampling on GenEval. \texttt{CoRe}$^2$ performs better than the standard sampling.}  
\vspace{-8pt}
\scalebox{0.8}{
\setlength{\tabcolsep}{12pt} 
\begin{tabular}{lccccccc}
\Xhline{3\arrayrulewidth}\\[-2ex]
{\textbf{Method}} & {\textbf{Single (↑)}} & {\textbf{Two (↑)}} & {\textbf{Counting (↑)}} & {\textbf{Colors (↑)}} & {\textbf{Positions (↑)}} & {\textbf{Color Attribution (↑)}} & {\textbf{Overall (↑)}}\\
\Xhline{3\arrayrulewidth}
        \\[-2ex]
        \multicolumn{7}{l}{\large\textit{\textbf{SDXL}}}\\\Xhline{3\arrayrulewidth}
Standard 
 & 97.50\% & 69.70\% & 35.00\% & 86.17\% & 13.00\% & \CC{25.00\%} & 54.39\% \\
\texttt{CoRe}$^2$ (Ours) 
 & \CC{97.50\%} & \CC{78.79\%} & \CC{42.50\%} & \CC{87.23\%} & \CC{14.00\%} & 16.00\% & \CC{56.01\%} \\[0ex]
\multicolumn{7}{l}{\large\textit{\textbf{SD3.5}}}\\\Xhline{3\arrayrulewidth}
Standard 
 & 100.00\% & 85.86\% & 72.50\% & \CC{81.91\%} & 22.00\% & 56.00\% & 69.71\% \\
\texttt{CoRe}$^2$ (Ours) 
 & \CC{100.00\%} & \CC{87.88\%} & \CC{73.75\%} & 79.79\% & \CC{27.00\%} & \CC{59.00\%} & \CC{71.23\%} \\
\Xhline{3\arrayrulewidth}\\[-2ex]
\end{tabular}
}
\label{tab:comparison_geneval}
\vspace{-10pt}
\end{table*}

%% file: tables/main_t2i_compbench.tex
\begin{table*}[!t]
\centering
\vskip -0.01in
\small
\caption{Comparison between \texttt{CoRe}$^2$ and the standard sampling on T2I-Compbench. \texttt{CoRe}$^2$ performs better than the standard sampling.}  
\vspace{-8pt}
\scalebox{0.72}{
\setlength{\tabcolsep}{12pt} 
\begin{tabular}{lccccccccc}
\Xhline{3\arrayrulewidth}\\[-2ex]
{\textbf{Method}} & \multicolumn{3}{c}{\textbf{Attribute Binding (↑)}} & \multicolumn{3}{c}{\textbf{Object Relationship (↑)}} & {\textbf{Complex (↑)}} & {\textbf{Numeracy (↑)}} & {\textbf{Overall (↑)}} \\
\cmidrule(lr){2-4} \cmidrule(lr){5-7}
 & Color & Shape & Texture & 2D-Spatial & 3D-Spatial & Non-Spatial &  &  & \\
\Xhline{3\arrayrulewidth}
        \\[-2ex]
        \multicolumn{9}{l}{\large\textit{\textbf{SDXL}}}\\\Xhline{3\arrayrulewidth}
Standard 
 & 0.5632 & 0.4746 & 0.5247 & \CC{0.2056} & 0.3577 & 0.3109 & 0.5120 & 0.3409 & 0.4112 \\
\texttt{CoRe}$^2$ (Ours) 
 & \CC{0.6000} & \CC{0.5194} & \CC{0.5543} & 0.1941 & \CC{0.3625} & \CC{0.3126} & \CC{0.5224} & \CC{0.3453} & \CC{0.4264} \\[0ex]

\multicolumn{9}{l}{\large\textit{\textbf{SD3.5}}}\\\Xhline{3\arrayrulewidth}
Standard 
 & 0.8036 & 0.5851 & 0.7166 & 0.2542 & 0.3942 & 0.3171 & \CC{0.6277} & 0.3780 & 0.5096 \\
\texttt{CoRe}$^2$ (Ours) 
 & \CC{0.8090} & \CC{0.5965} & \CC{0.7245} & \CC{0.2733} & \CC{0.3953} & \CC{0.3177} & 0.6169 & \CC{0.3862} & \CC{0.5150} \\
\Xhline{3\arrayrulewidth}\\[-2ex]
\end{tabular}
}
\label{tab:comparison_t2i_compbench}
\vspace{-18pt}
\end{table*}

%% file: sec/appendix.tex
\appendix
\clearpage
\onecolumn

\section*{\Large{Appendix}}

\section{Benchmark and Dataset}
\label{apd:benchmark_and_dataset}

In this section, we provide an overview of the benchmarks, evaluation metrics, and the datasets obtained in the \textit{Collect} phase used in our main paper.

\subsection{Benchmark}

\paragraph{Pick-a-Pic.} Pick-a-Pic~\citep{pick_of_pic} is an open dataset curated to capture user preferences for T2I synthesized images. The dataset is collected via an intuitive web application, allowing users to synthesize images and express their preferences. The Pick-a-Pic dataset includes over 500,000 examples spanning 35,000 unique prompts.

\paragraph{DrawBench.} DrawBench\footnote{https://huggingface.co/datasets/shunk031/DrawBench} is a benchmark dataset introduced to enable comprehensive evaluation of text-to-image synthesis models. It consists of 200 meticulously designed prompts, organized into 11 categories to assess model capabilities across a range of semantic dimensions. These dimensions include compositionality, numerical reasoning, spatial relationships, and the ability to interpret complex textual instructions. DrawBench is specifically designed to provide a multidimensional analysis of model performance, facilitating the identification of both strengths and weaknesses in text-to-image synthesis.

\paragraph{HPD v2.} The human preference dataset v2 (HPD v2)~\citep{HPSV2} is an extensive dataset featuring clean and precise annotations, designed to capture user preferences for images generated from text prompts. It includes 798,090 binary preference labels across 433,760 image pairs, aiming to address the limitations of conventional evaluation metrics that fail to accurately reflect human preferences. Following the methodologies presented in~\citep{HPSV2,meissonic}, we employed four distinct subsets for our analysis: Animation, Concept-art, Painting, and Photo, each containing 800 prompts.

\paragraph{GenEval.} GenEval is an evaluation framework specifically tailored to assess the compositional properties of generated images, such as object co-occurrence, spatial positioning, object count, and color. By leveraging SOTA detection models, GenEval provides a robust evaluation of text-to-image generation tasks, ensuring strong alignment with human judgments. Additionally, the framework allows for the integration of other advanced vision models to validate specific attributes, such as object color. The benchmark comprises 550 prompts, all of which are straightforward and easy to interpret.

\paragraph{T2I-Compbench.} T2I-Compbench is a comprehensive benchmark designed to evaluate open-world compositional text-to-image synthesis. It includes 6,000 compositional text prompts, systematically categorized into three primary groups: attribute binding, object relationships, and complex compositions. These groups are further divided into six subcategories: color binding, shape binding, texture binding, spatial relationships, non-spatial relationships, and intricate compositions.

\subsection{Evaluation Metric}

\paragraph{PickScore.} PickScore is a CLIP-based scoring model, developed using the Pick-a-Pic dataset, which captures user preferences for generated images. This metric demonstrates performance surpassing that of typical human benchmarks in predicting user preferences. By aligning effectively with human evaluations and leveraging the diverse range of prompts in the Pick-a-Pic dataset, PickScore offers a more relevant and insightful assessment of text-to-image models compared to traditional metrics like FID~\citep{fid} on datasets such as MS-COCO~\citep{COCO}.

\paragraph{HPS v2.} The human preference score version 2 (HPS v2) is an improved model to predict user preferences, achieved by fine-tuning the CLIP model~\citep{CLIP} on the HPD v2. This refined metric is designed to align text-to-image generation outputs with human tastes by estimating the likelihood that a generated image will be preferred, thereby serving as a reliable benchmark for evaluating the performance of text-to-image models across diverse image distributions.

\paragraph{AES.} The Aesthetic Score (AES)~\citep{AES} is calculated using a model built on CLIP embeddings, further enhanced with multilayer perceptron (MLP) layers to evaluate the visual appeal of images. This metric provides a quantitative measure of the aesthetic quality of generated images, offering valuable insights into their alignment with human aesthetic standards.

\paragraph{ImageReward.} ImageReward~\citep{Imagereward} is a specialized reward model designed to evaluate text-to-image synthesis through human preferences. Trained on an extensive dataset of human comparisons, the model effectively captures user inclinations by assessing multiple aspects of synthesized images, including their alignment with text prompts and their aesthetic quality. ImageReward has shown superior performance compared to traditional metrics such as the Inception Score (IS)~\citep{is} and Fréchet Inception Distance (FID), making it a highly promising tool for automated evaluation in text-to-image synthesis tasks.

\subsection{Dataset}

\paragraph{SD3.5.} For SD3.5, we initially collected a total of 200k prompts to build a dataset based on SD3.5. Among these, 100k prompts were sourced from Pick-of-Pic, refined using InternVL2-26b to ensure high-quality prompts. The remaining 100k prompts were obtained from the diffusiondb-prompt-upscale\footnote{https://huggingface.co/datasets/adams-story/diffusiondb-prompt-upscale} dataset, which were also refined using InternVL2-26b. Using these refined prompts, we were able to generate 100k CFG trajectories.

\paragraph{SDXL.} The prompt used on SDXL was the same as SD3.5, but we simply randomly selected 100k of them to generate CFG trajectories.

\paragraph{LlamaGen.} In LlamaGen, we employ the same prompts as SDXL, utilizing 100k prompts to construct its dataset. Furthermore, to ensure enhanced generative performance, our implementation of LlamaGen differs from the official version. Specifically, we adjusted the CFG scale to 9 and incorporated a comprehensive negative prompt instead of using a null prompt. The negative prompt includes: ``worst quality, normal quality, low quality, low res, blurry, distortion, text, watermark, logo, banner, extra digits, cropped, jpeg artifacts, signature, username, error, sketch, duplicate, ugly, monochrome, horror, geometry, mutation, disgusting, bad anatomy, bad proportions, bad quality, deformed, disconnected limbs, out of frame, out of focus, dehydrated, disfigured, extra arms, extra limbs, extra hands, fused fingers, gross proportions, long neck, jpeg, malformed limbs, mutated, mutated hands, mutated limbs, missing arms, missing fingers, picture frame, poorly drawn hands, poorly drawn face, collage, pixel, pixelated, grainy, color aberration, amputee, autograph, bad illustration, beyond the borders, blank background, body out of frame, boring background, branding, cut off, dismembered, disproportioned, distorted, draft, duplicated features, extra fingers, extra legs, fault, flaw, grains, hazy, identifying mark, improper scale, incorrect physiology, incorrect ratio, indistinct, kitsch, low resolution''.

\section{Detail Implementation}
\label{apd:detail_implementation}

\subsection{Architecture of Noise Model}

Here, we provide a detail description of the noise model architecture used in SD3.5, SDXL, and LlamaGen. Given that the open-source versions of FLUX are distilled through CFG distillation technique, we propose a modified \texttt{CoRe}$^2$ to enhance the model's performance on FLUX. Further details can be found in Appendix~\ref{apd:flux_experiment}. 

\paragraph{SD3.5.} For SD3.5's noise model, we utilize the MM-DiT-Block architecture with an input latent shape of 128$\times$128. The patch size is set to 2$\times$2, and the input and output latent channel sizes are both 16. The architecture employs 8 MM-DiT-Blocks, with 24 attention heads incorporated. Text embedding information is injected following the structure of DiT, ensuring seamless integration of textual data into the model.

\paragraph{SDXL.} For SDXL's noise model, we adopt the MM-DiT-Block architecture with an input latent shape of 128$\times$128. The patch size is configured as 2$\times$2, while the input and output latent channel sizes are both set to 4. The architecture includes 4 MM-DiT-Blocks, with 24 attention heads utilized. Text embedding information is incorporated following the DiT structure, enabling smooth and effective integration of textual data into the model.

\paragraph{LlamaGen.} Since both the input and output of LlamaGen are logits, the MM-DiT-Block cannot be used to construct the noise model. To reduce the gap between LlamaGen and the noise model, we decided to use Llama blocks as the components for the noise model. Furthermore, while the number of layers was reduced from LlamaGen's 36 layers to 12 layers, all other settings remain unchanged. For the input and output of the noise model, as they are entirely in the form of logits, the ``tok\_embedding'' layer, which typically converts discrete vectors into continuous encodings, is unnecessary and therefore omitted.

\subsection{Experimental Setup of the \textbf{\textit{Reflect}} Phase}

For the training of noise models across SDXL, SD3.5, and LlamaGen, we adopt a per-GPU batch size of 4, combined with gradient accumulation over 2 steps. This setup produces a global batch size of 64 when utilizing 8 GPUs. For SDXL and SD3.5, we rely on the mean squared error (MSE) loss as the default choice for CFG distillation. In contrast, for LlamaGen, we use MSE loss during the initial training phase and later transition to Kullback-Leibler (KL) divergence, aiming to emphasize the ``easy-to-learn'' components. The learning rate is set to 2e-6 for SDXL and SD3.5, while LlamaGen operates with a higher learning rate of 1e-4. All models utilize the AdamW optimizer with a weight decay of 1e-3, and training is conducted using a cosine learning rate scheduler to ensure smooth convergence.

For the number of training iterations, we conduct ablation experiments on SD3.5 and found that among 1k, 10k, and 50k iterations, the noise model achieve the best performance with 10k iterations. Therefore, we adopt this setting for both SDXL and LlamaGen as well.

\subsection{Experimental Setup of the \textbf{\textit{Refine}} Phase}

During the \textit{Refine} stage, we adhere to the default sampling steps for both SD3.5 and SDXL, which are set to 28 and 50, respectively. Additionally, our ablation studies revealed that $\omega_\textrm{w2s}$ achieves optimal performance with values of 2.5 on SD3.5 and 1.5 on SDXL. Unless otherwise specified, the experiments presented in our tables were all conducted using the full path executed in slow mode.

\section{How Does W2S Guidance Work?}
\label{apd:do_w2s_work}

Here, we provide a detailed theoretical analysis and proof of why W2S guidance works effectively in DMs. It is worth noting that this proof assumes a premise where the weak model and the strong model perform comparably in learning easily learnable content, but exhibit a significant gap in their ability to capture the more difficult-to-learn content.

\begin{lemma}
\label{llama:1}
    Assuming \( p(x_0) \) can be modeled as mixture Gaussian model (GMM) where each Gaussian component is a Dirac distribution, i.e., \( p(x_0) = \sum_{i=1}^M w_i \delta(x - x_i) \) ($M$ is the number of Gaussian components), it follows that both \( p(x_t) \) and \( p(x_s|x_t) \) are Gaussian distributions. Furthermore, the score function estimation \( \epsilon_\theta(x_t, t) \) can be expressed as a summation of the logarithms of the Gaussian components.
\end{lemma}

We know $p(x_s|x_t) = \int p(x_s|x_t,x_0)p(x_0|x_t)dx_0$, where $q(x_0)= \sum_{i=1}^M w_i \delta(x - x_i)$ subject to $\sum_{i=1}^{M} w_i = 1$. Then, we can get\begin{equation}
    \begin{split}
        p(x_s|x_t)&=
\int p(x_s|x_t,x_0)p(x_0|x_t)dx_0\\
& = \int p(x_s|x_t,x_0)p(x_0)p(x_t|x_0)/p(x_t)dx_0 \\
&=1/q(x_t) \int \sum_{i=1}^Mp(x_s|x^i_t,x^i_0)w_ip(x^i_t|x^i_0).\\
    \end{split}
\end{equation}
Since the forward process and the reverse process can be described as $p(x_t|x_0) = \mathcal{N}(x_t|\alpha_tx_0, \sigma_t^2)$ and $p(x_s|x_t,x_0) =\mathcal{N}_\theta(x_0|\alpha_{s|t}x_t, -\alpha_{s|t}^2[\sigma_t-\sigma_s]^2)$, where $\mathcal{N}_\theta$ in the second term stands for the predicted noise obtained by the score function estimator. In the DDIM process, the reverse process can also be formulated as\begin{equation}
    \begin{split}
        x_s &= \alpha_s \frac{x_t - \sigma_t \epsilon_\theta(x_t, t)}{\alpha_t} + \sigma_s \epsilon_\theta(x_t, t).\\
    \end{split}
\end{equation}
Since $\epsilon_\theta(x_t,t)$ itself is fitting the score function, we have 
\begin{equation}
    \begin{split}
        \frac{\epsilon_\theta(x_t,t)}{-\sigma_t}  &= \nabla_{x_t}\log p_t(x_t) \propto  \nabla_{x_t}\sum_{i=1}^P w_i\log \mathcal{N}(\mu_i(x),\Sigma_i(x)).
    \end{split}
\end{equation}

The condition we know is \( p(x_0) \) is a GMM, and \( p(x_t) \) can be derived as:  
\begin{equation}
    \begin{split}
p(x_t) &= \int p(x_t|x_0)p(x_0)dx_0 \\
&= \int \sum_{i=1}^M w_i \frac{1}{\sqrt{2\pi}\sigma_t} \exp\left(-\frac{\Vert x_t - \alpha_t x_0 \Vert_2^2}{2\sigma_t^2}\right) dx_0.\\
    \end{split}
\end{equation}

This simplifies to:  
\begin{equation}
    \begin{split}
p(x_t) &= \frac{1}{\sqrt{2\pi}\sigma_t} \exp\left(-\frac{\Vert x_t - \alpha_t x_0 \Vert_2^2}{2\sigma_t^2}\right)\\
&= \mathcal{N}(x_t | \alpha_t x_0, \sigma_t^2).\\
    \end{split}
\end{equation}
So $p(x_t)$ is a Gaussian distribution, and we can also get $p(x_s|x_t) = \mathcal{N}(x_s|\frac{\alpha_{t|s}\sigma_s^2}{\alpha_t^2}x_t+\frac{\alpha_s\sigma_{t|s}^2}{\alpha_t^2}x_0,\frac{\alpha^2_s\sigma_{t|s}^2}{\alpha_t^2})$. Therefore, modeling the predicted noise as $\frac{\epsilon_\theta(x_t, t)}{-\sigma_t} = \nabla_{x_t} \log p_t(x_t) \propto \nabla_{x_t} \sum_{i=1}^P w_i \log \mathcal{N}(\mu_i(x), \Sigma_i(x))$ is reasonable. This term remains a Gaussian distribution, where the Gaussian components—if the logarithm is removed—form a GMM.

\begin{definition}
\label{apd:def1}
Let \( g(x) = \sum_{i=1}^M w_i \mathcal{N}(\mu_i, \Sigma_i) \) be the ground truth GMM with \( \sum_{i=1}^M w_i = 1 \). We partition the components into \textbf{easy-to-learn} and \textbf{difficult-to-learn} subsets based on their approximation errors under a fitted GMM \( \tilde{g}(x) = \sum_{i=1}^M w_i \mathcal{N}(\tilde{\mu}_i, \tilde{\Sigma}_i) \):  

\paragraph{Easy-to-Learn Components (\( i \in \{1,\ldots,M_1\} \)).} The approximation error for these components satisfies:  
\begin{equation}
    \begin{split}
       & \mathbb{E}_{i \in [1,M_1]}\left[D_{\text{KL}}\left(\mathcal{N}(\mu_i, \Sigma_i) \Vert \mathcal{N}(\tilde{\mu}_i, \tilde{\Sigma}_i)\right)\right] \leq \eta_{\text{easy}},\\   
    \end{split}
\end{equation}
where \( D_{\text{KL}} \) is the KL divergence. Under the assumption \( \tilde{\Sigma}_i = \Sigma_i \), this simplifies to  
\begin{equation}
    \begin{split}
   &\sum_{i=1}^{M_1} w_i \|\mu_i - \tilde{\mu}_i\|^2 \leq \eta_{\text{easy}}.\\
    \end{split}
\end{equation}

\paragraph{Difficult-to-Learn Components (\( i \in \{M_1+1,\ldots,M\} \)).} These components exhibit larger approximation errors:  
\begin{equation}
    \begin{split}
   \eta_{\text{easy}} &< \sum_{i=M_1+1}^M w_i \|\mu_i - \tilde{\mu}_i\|^2 \leq \eta_{\text{difficult}}.\\
    \end{split}
\end{equation}

\end{definition}

Based on the above lemma and definition, we can get the following theorem:

\begin{theorem}

Let us assume there are two macro-level diffusion models, denoted as \( \epsilon_\theta^\textrm{weak} \) and \( \epsilon_\theta^\textrm{strong} \). According to Definition~\ref{apd:def1}, these models handle ``easy-to-learn'' and ``difficult-to-learn'' content, where it is assumed that the contents overlap. The key distinction (constraint) is as follows:  
\begin{equation}
    \begin{split}
        & |\eta_\textrm{easy}^\textrm{weak} - \eta_\textrm{easy}^\textrm{strong}| < |\eta_\textrm{difficult}^\textrm{weak} - \eta_\textrm{difficult}^\textrm{strong}|, \\
        & \eta_\textrm{difficult}^\textrm{strong} < \eta_\textrm{difficult}^\textrm{weak}, \quad \eta_\textrm{easy}^\textrm{strong} < \eta_\textrm{easy}^\textrm{weak},\\
&(\mu_i - \mu_i^\textrm{strong})(\mu_i^\textrm{strong} - \mu_i^\textrm{weak}) > 0.\\
    \end{split}
\end{equation} There exists $\omega_\textrm{w2s} > 1$ (i.e., the W2S guidance scale) such that W2S guidance $\epsilon_\theta^\textrm{weak}(x_t,t) + \omega_\textrm{w2s}\left[\epsilon_\theta^\textrm{strong}(x_t,t)- \epsilon_\theta^\textrm{weak}(x_t,t)\right]$ is closer to the ground truth distribution $\sum_{i=1}^M w_i \mathcal{N}(\mu_i, \Sigma_i)$ than either $\epsilon_\theta^\textrm{weak}(x_t,t)$ or $\epsilon_\theta^\textrm{strong}(x_t,t)$ alone.
\end{theorem}

\begin{proof}
Note that $\epsilon_\theta^\textrm{weak}(x_t,t) + \omega_\textrm{w2s}\left[\epsilon_\theta^\textrm{strong}(x_t,t)- \epsilon_\theta^\textrm{weak}(x_t,t)\right]$ can be rewritten as (use Lemma~\ref{llama:1})
\begin{equation}
    \begin{split}
        &\frac{1}{-\sigma_t}\nabla_{x_t}\left[\sum_{i=1}^M w_i\left[\log \mathcal{N}(\mu^\textrm{strong}_i(x),\Sigma^\textrm{strong}_i(x))^{\omega_\textrm{w2s}}+\log \mathcal{N}(\mu^\textrm{weak}_i(x),\Sigma^\textrm{weak}_i(x))^{(1-\omega_\textrm{w2s})}\right]\right].\\
    \end{split}
\end{equation}
Each component can be further derived as
\begin{equation}
    \begin{split}
        & \log \mathcal{N}(\mu^\textrm{strong}_i(x),\Sigma^\textrm{strong}_i(x))^{\omega_\textrm{w2s}}+ \log \mathcal{N}(\mu^\textrm{weak}_i(x),\Sigma^\textrm{weak}_i(x))^{(1-\omega_\textrm{w2s})}\\
        & \propto \log(\exp(-\frac{\omega_\textrm{w2s}\Vert x - \mu_i^\textrm{strong}(x)\Vert_2^2}{2\Sigma^\textrm{strong}_i(x)} +\frac{(\omega_\textrm{w2s}-1)\Vert x - \mu_i^\textrm{weak}(x)\Vert_2^2}{2\Sigma^\textrm{weak}_i(x)} )) \\
        & = \frac{-\Sigma^\textrm{weak}_i(x)\omega_\textrm{w2s}(x^Tx-2x\mu_i^\textrm{strong}(x)+\mu_i^{\textrm{strong},2}(x))+\Sigma^\textrm{strong}_i(x)(\omega_\textrm{w2s}-1)(x^Tx-2x\mu_i^\textrm{weak}(x)+\mu_i^{\textrm{weak},2}(x))}{4\Sigma^\textrm{strong}_i(x)\Sigma^\textrm{weak}_i(x)} \\
        & =\frac{\kappa}{4\Sigma^\textrm{strong}_i(x)\Sigma^\textrm{weak}_i(x)} \left\| x + \frac{\Sigma^\textrm{weak}_i(x)\omega_\textrm{w2s} \mu_i^\textrm{strong}(x) - \Sigma^\textrm{strong}_i(x)(\omega_\textrm{w2s}-1)\mu_i^\textrm{weak}(x)}{\kappa} \right\|_2^2 + C, \\
    \end{split}
\end{equation}
where $\mu_i^\text{strong}(x)$, $\mu_i^\text{weak}(x)$, $\Sigma_i^\text{strong}(x)$, $\Sigma_i^\text{weak}(x)$ stand for the expectation of the strong model, the expectation of the weak model, the variance of the strong model, the variance of the weak model, respectively, and $C$ refer to a constant. We aim to prove the existence of a weighting factor $\omega_\textrm{w2s}$ such that the combined mean $B(\omega_\textrm{w2s}) = -\frac{\Sigma^\textrm{weak}_i \omega_\textrm{w2s} \mu_i^\textrm{strong} - \Sigma^\textrm{strong}_i (\omega_\textrm{w2s}-1)\mu_i^\textrm{weak}}{\kappa}, \quad \kappa = -\Sigma^\textrm{weak}_i \omega_\textrm{w2s} + \Sigma^\textrm{strong}_i (\omega_\textrm{w2s}-1)$ is closer to the target mean $\mu_i$ than either $\mu_i^\textrm{weak}$ or $\mu_i^\textrm{strong}$ alone. This is equivalent to minimizing the squared distance:

\begin{equation}
    \begin{split}
        & E(\omega_\textrm{w2s}) = \left\| B(\omega_\textrm{w2s}) - \mu_i \right\|^2_2. \\
    \end{split}
\end{equation}

First, we should simplify $B(\omega_\textrm{w2s})$. Assume $\Sigma^\textrm{weak}_i = \Sigma^\textrm{strong}_i = \Sigma_i$ for simplicity. Then $B(\omega_\textrm{w2s}) = \omega_\textrm{w2s} \mu_i^\textrm{strong} + (1-\omega_\textrm{w2s})\mu_i^\textrm{weak}$. Second, we need to give the error function $E(\omega_\textrm{w2s}) = \left\| \omega_\textrm{w2s} (\mu_i^\textrm{strong} - \mu_i^\textrm{weak}) + (\mu_i^\textrm{weak} - \mu_i) \right\|^2$. Review the conditions given by the theorem, we know: \textit{\textcolor{C3}{1)}} $\eta^\textrm{strong}_\textrm{difficult} < \eta^\textrm{weak}_\textrm{difficult}$, implying $\|\mu_i^\textrm{strong} - \mu_i\| < \|\mu_i^\textrm{weak} - \mu_i\|$, and \textit{\textcolor{C3}{2)}} $(\mu_i - \mu_i^\textrm{strong})(\mu_i^\textrm{strong} - \mu_i^\textrm{weak}) > 0$, meaning $\mu_i^\textrm{strong}$ lies between $\mu_i^\textrm{weak}$ and $\mu_i$. Third, we can give the derivation of $E(\omega_\textrm{w2s})$ with respect to $\omega_\textrm{w2s}$ as $E'(\omega_\textrm{w2s}) = 2(\mu_i^\textrm{strong} - \mu_i^\textrm{weak}) \cdot \left( \omega_\textrm{w2s} (\mu_i^\textrm{strong} - \mu_i^\textrm{weak}) + (\mu_i^\textrm{weak} - \mu_i) \right)$.

\begin{enumerate}
    \item At $\omega_\textrm{w2s} = 0$, $E'(0) = 2(\mu_i^\textrm{strong} - \mu_i^\textrm{weak})(\mu_i^\textrm{weak} - \mu_i) < 0$ (since $\mu_i^\textrm{strong}$ is closer to $\mu_i$).
    \item At $\omega_\textrm{w2s} = 1$, $E'(1) = 2(\mu_i^\textrm{strong} - \mu_i^\textrm{weak})(\mu_i^\textrm{strong} - \mu_i) < 0$ (by the same reasoning).
\end{enumerate}

Finally, we optimize \( \omega_\textrm{w2s} \). Since \( E'(\omega_\textrm{w2s}) \) is linear in \( \omega_\textrm{w2s} \), and \( E'(0) < 0 \), \( E'(1) < 0 \), the minimum of \( E(\omega_\textrm{w2s}) \) occurs at \( \omega_\textrm{w2s} > 1 \). Specifically, solving \( E'(\omega_\textrm{w2s}) = 0 \) gives:  
\begin{equation}
    \begin{split}
        \omega_\textrm{w2s}^* &= \frac{\mu_i - \mu_i^\textrm{weak}}{\mu_i^\textrm{strong} - \mu_i^\textrm{weak}}.\\
    \end{split}
\end{equation}
Given the condition \( (\mu_i - \mu_i^\textrm{strong})(\mu_i^\textrm{strong} - \mu_i^\textrm{weak}) > 0 \), it follows that \( \omega_\textrm{w2s}^* > 1 \), which ensures \( E(\omega_\textrm{w2s}^*) < \eta^\textrm{strong}_\textrm{difficult} \). Thus, there exists \( \omega_\textrm{w2s} > 1 \) such that the mean square error (MSE) \(\Vert B(\omega_\textrm{w2s}) - \mu_i \Vert_2^2 \) is smaller than either \( \Vert \mu_i^\textrm{weak}- \mu_i \Vert_2^2 \) or \( \Vert \mu_i^\textrm{strong}- \mu_i \Vert_2^2 \) individually. This concludes the proof.
\end{proof}

\subsection{Experiment on FLUX}
\label{apd:flux_experiment}
\begin{figure*}
    \centering
    \includegraphics[width=.9\linewidth]{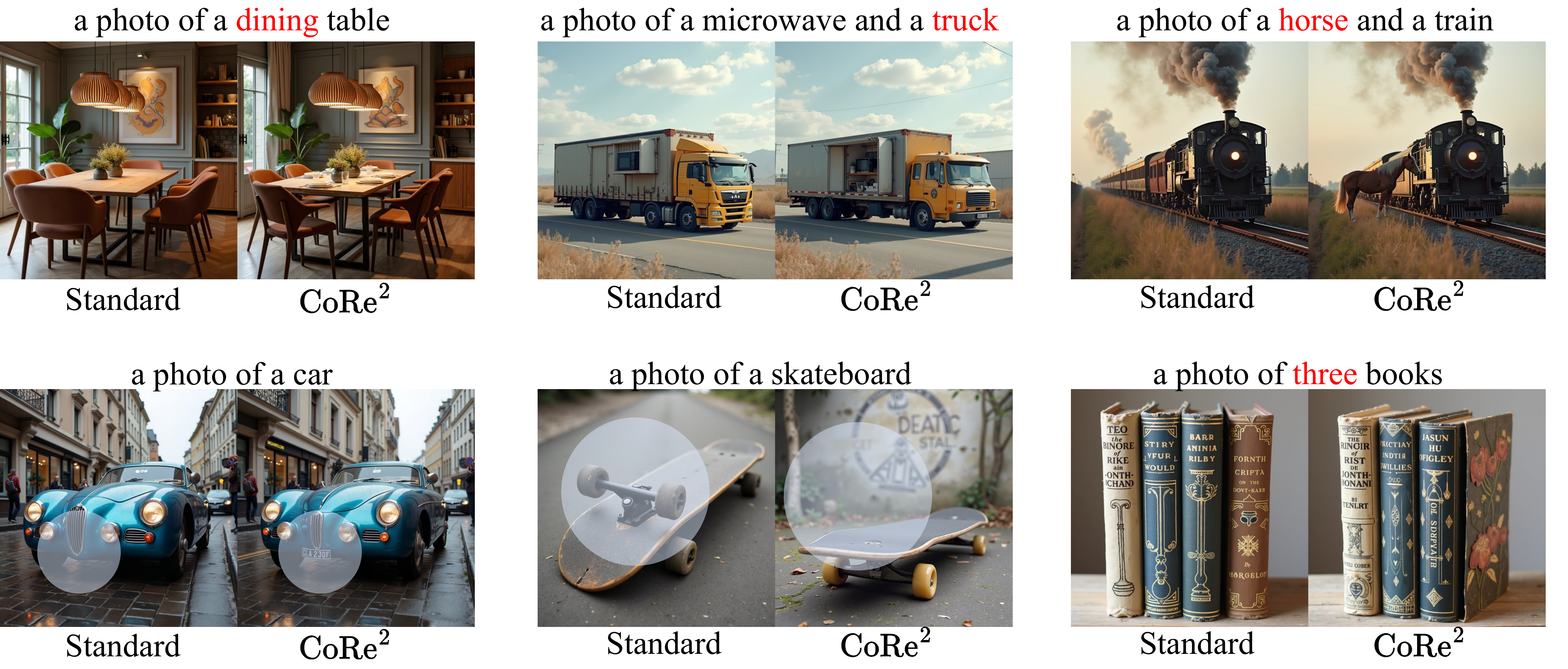}
    \vspace{-10pt}\caption{Visualization of the (modified) \texttt{CoRe}$^2$ and the standard sampling on FLUX. Compared to \texttt{CoRe}$^2$, which is obtained through the complete three-stage pipeline, the (modified) \texttt{CoRe}$^2$ faces a limitation due to the lack of an open-sourced version of FLUX without CFG distillation. As a result, we treat FLUX.1-schnell a weak model, while FLUX.1-dev as a strong model when executing W2S guidance. Although FLUX falls slightly short in delivering the significant gains over standard sampling as seen in \texttt{CoRe}$^2$ in Fig.~\ref{fig:main_visualization}, it nonetheless demonstrates a certain degree of improvement in both visual quality and semantic consistency of the generated images.}
    \label{fig:flux_presentation}
    \vspace{-10pt}
\end{figure*}
\input{tables/apd_flux_pick}
\input{tables/apd_flux_draw}
\input{tables/apd_flux_geneval}
\input{tables/apd_flux_hpd}
Given that Black Forest Lab has only open-sourced FLUX.1-dev and FLUX.1-schnell, both of which are derived using CFG distillation, it becomes impractical to employ the \textit{Collect}, \textit{Reflect}, and \textit{Refine} stages from the main paper to enhance the high-frequency details in the images. Instead, we opted for a more flexible implementation that marginally lowers the performance upper bound of \texttt{CoRe}$^2$. Specifically, FLUX.1-dev is designated as the strong model, while FLUX.1-schnell acts as the weak model. Interestingly, experimental results reveal that FLUX.1-schnell is not necessarily weaker than FLUX.1-dev, which results in the optimal $\omega_\textrm{w2s}$ value potentially falling between 0 and 1. Through relevant ablation experiments, as presented in Table~\ref{tab:pic_of_pick}, we determine that $\omega_\textrm{w2s} = 0.85$ yields the best performance. 

Furthermore, we conduct comparative evaluations between \texttt{CoRe}$^2$ and the standard sampling on both GenEval and DrawBench benchmarks with FLUX.1-dev in Table~\ref{tab:drawbench} and Table~\ref{tab:geneval}, respectively. The outcomes demonstrate that \texttt{CoRe}$^2$ significantly enhances the fidelity of the generated images. Lastly, the visualization provided in Fig.~\ref{fig:flux_presentation} highlights that even this adapted version of \texttt{CoRe}$^2$ delivers a noticeable improvement in image details that is appreciable to the naked eye.

\begin{figure*}
    \centering
    \includegraphics[width=1.0\linewidth]{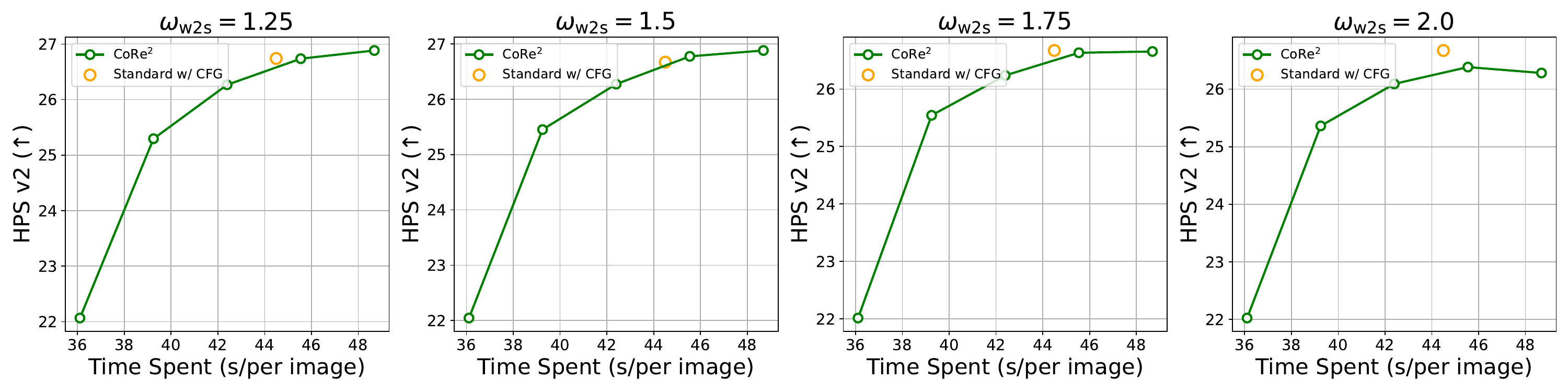}
    \vspace{-23pt}
    \caption{Ablation study of \( \omega_\textrm{w2s} \) on LlamaGen. We observe that \texttt{CoRe}$^2$ achieves optimal performance when \( \omega_\textrm{w2s} = 1.5 \).}
    \label{fig:ablation_study_llamagen}
        \vspace{-12pt}
\end{figure*}

\subsection{W2S Guidance Scale $\omega_\textrm{w2s}$ on LlamaGen}
\label{apd:w2s_llamagen_ablation_study}

In this section, we present the findings of the ablation experiments on $\omega_\textrm{w2s}$ for LlamaGen, as depicted in Fig.~\ref{fig:ablation_study_llamagen}. Compared to SDXL and SD3.5, we notice that the tuning process for $\omega_\textrm{w2s}$ on LlamaGen is notably more delicate and requires heightened precision. To tackle this, a grid search with a finer interval of 0.25 is utilized to pinpoint the optimal value of $\omega_\textrm{w2s}$ that enables \texttt{CoRe}$^2$ to deliver its peak performance. An example of the grid search methodology is provided for demonstration. From Fig.~\ref{fig:ablation_study_llamagen}, it can be inferred that selecting $\omega_\textrm{w2s}$ as 1.5, particularly when the entire sampling path operates in slow mode, yields a significant performance boost compared to the standard sampling.

\subsection{Additional Visualization}
\label{apd:additional_visualization}

\begin{figure*}[h]
    \centering
    \includegraphics[width=.9\linewidth]{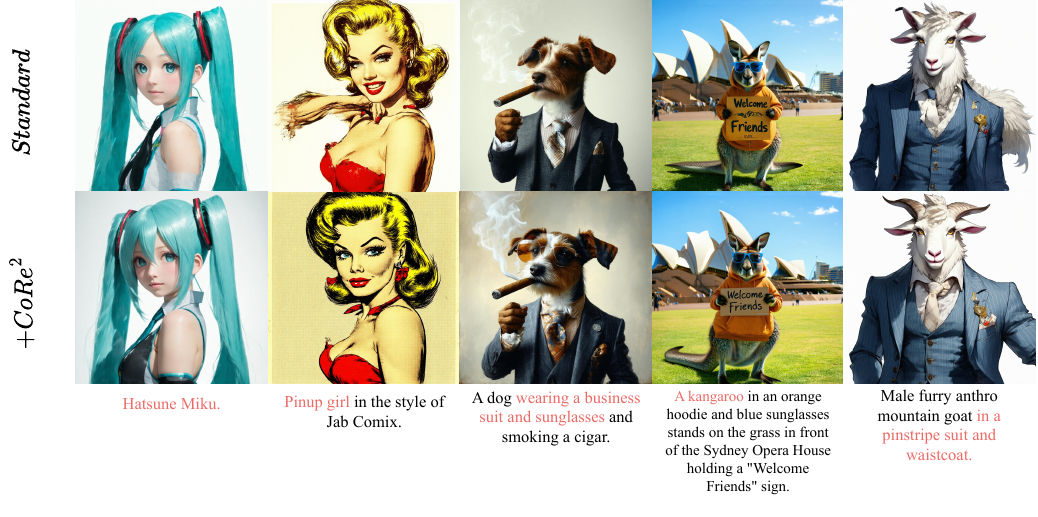}
    \vspace{-10pt}\caption{Visualization comparison about the standard sampling and \texttt{CoRe}$^2$ in anime style on SD3.5.}
    \label{fig:anime-sd35}
    \vspace{-10pt}
\end{figure*}

\begin{figure*}
    \centering
    \includegraphics[width=.9\linewidth]{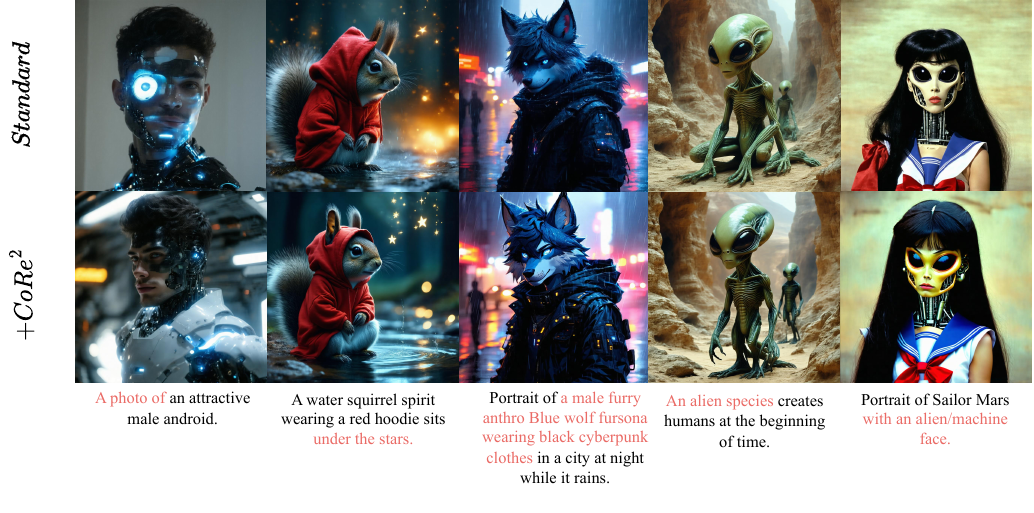}
    \vspace{-10pt}\caption{Visualization comparison about the standard sampling and \texttt{CoRe}$^2$ in concept-art style on SD3.5.}
    \label{fig:concept-sd35}
    \vspace{-10pt}
\end{figure*}

\begin{figure*}
    \centering
    \includegraphics[width=.9\linewidth]{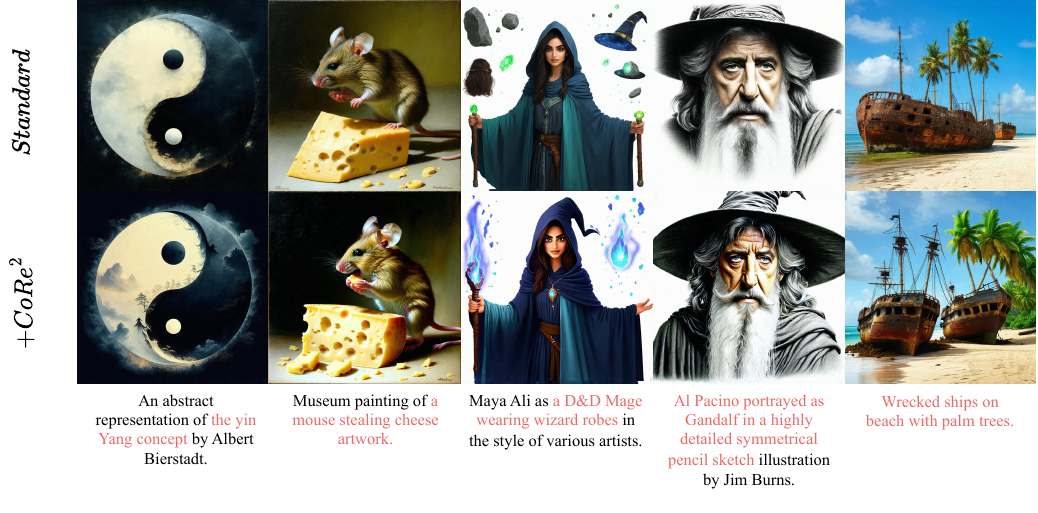}
    \vspace{-10pt}\caption{Visualization comparison about the standard sampling and \texttt{CoRe}$^2$ in painting style on SD3.5.}
    \label{fig:paint-sd35}
    \vspace{-10pt}
\end{figure*}

\begin{figure*}
    \centering
    \includegraphics[width=.9\linewidth]{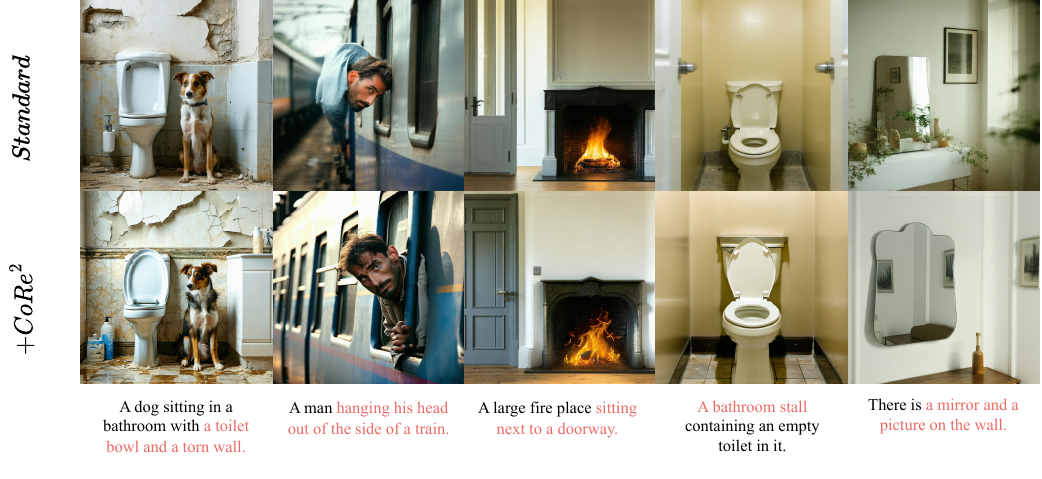}
    \vspace{-10pt}\caption{Visualization comparison about the standard sampling and \texttt{CoRe}$^2$ in photography style on SD3.5.}
    \label{fig:photo-sd35}
    \vspace{-10pt}
\end{figure*}

\begin{figure*}
    \centering
    \includegraphics[width=.9\linewidth]{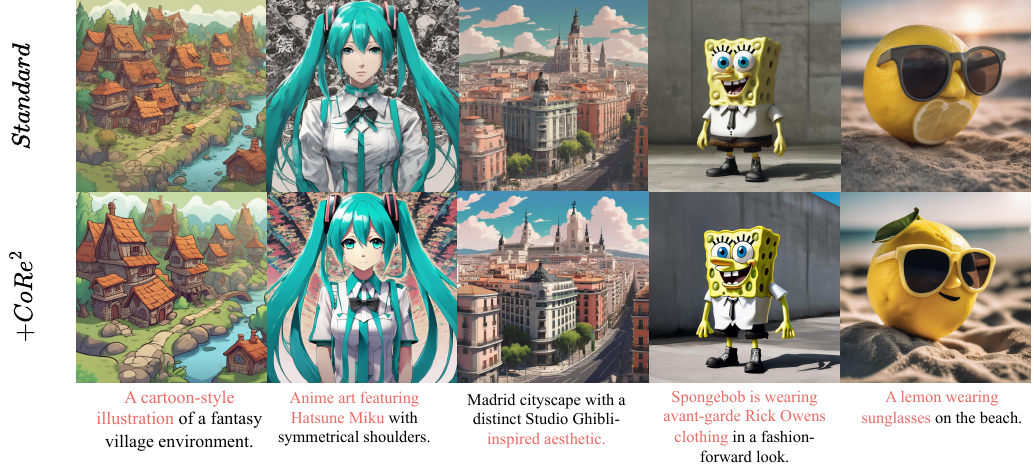}
    \vspace{-10pt}\caption{Visualization comparison about the standard sampling and \texttt{CoRe}$^2$ in anime style on SDXL.}
    \label{fig:anime-sdxl}
    \vspace{-10pt}
\end{figure*}

\begin{figure*}
    \centering
    \includegraphics[width=.9\linewidth]{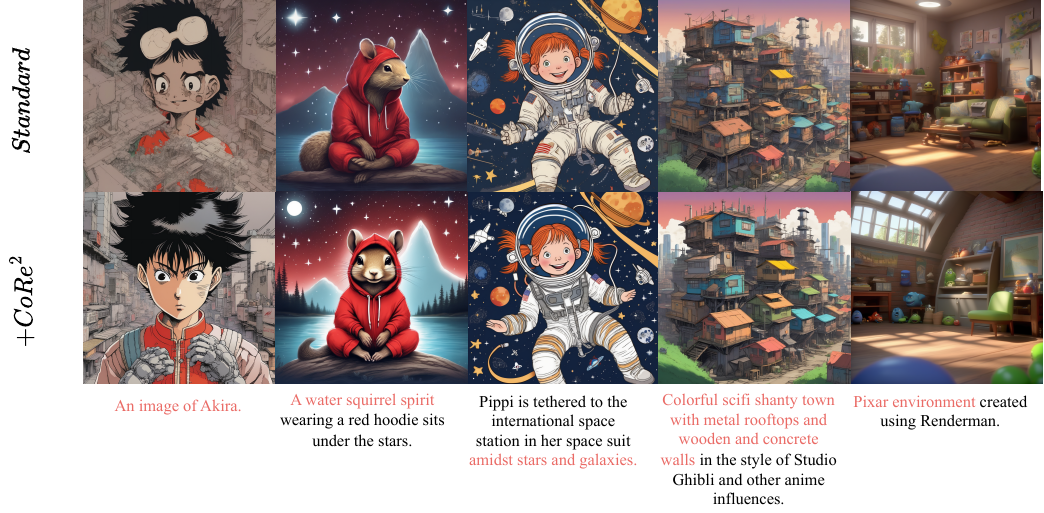}
    \vspace{-10pt}\caption{Visualization comparison about the standard sampling and \texttt{CoRe}$^2$ in concept-art style on SDXL.}
    \label{fig:concept-sdxl}
    \vspace{-10pt}
\end{figure*}

\begin{figure*}
    \centering
    \includegraphics[width=.9\linewidth]{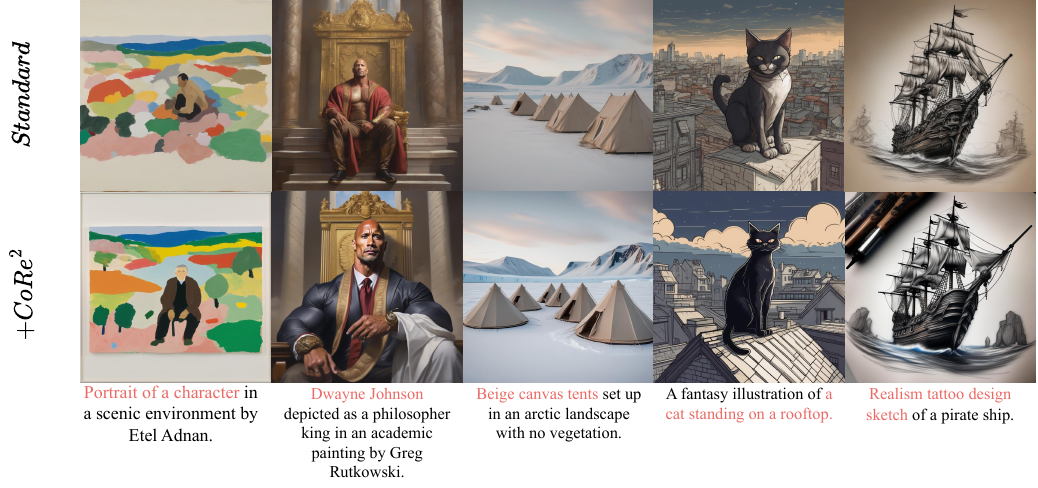}
    \vspace{-10pt}\caption{Visualization comparison about the standard sampling and \texttt{CoRe}$^2$ in painting style on SDXL.}
    \label{fig:paint-sdxl}
    \vspace{-10pt}
\end{figure*}

\begin{figure*}
    \centering
    \includegraphics[width=.9\linewidth]{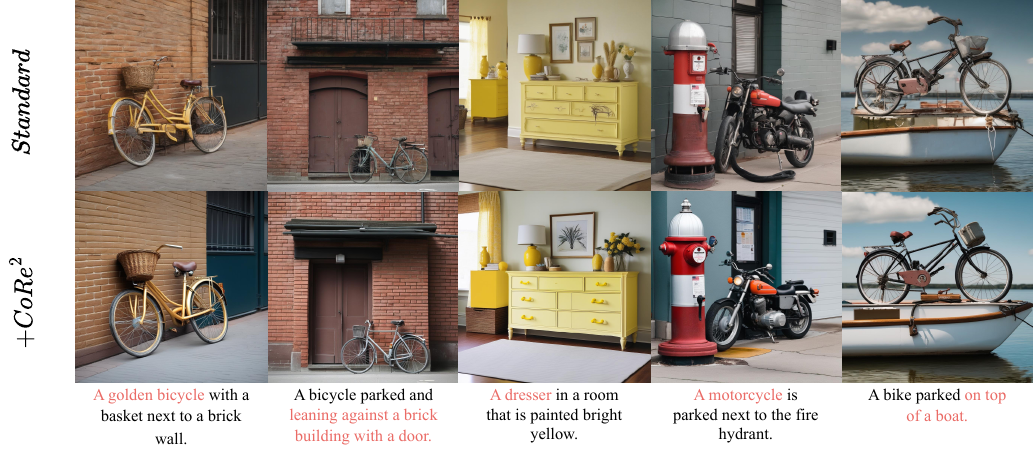}
    \vspace{-10pt}\caption{Visualization comparison about the standard sampling and \texttt{CoRe}$^2$ in photography style on SDXL.}
    \label{fig:photo-sdxl}
    \vspace{-10pt}
\end{figure*}

Here, we provide additional visualizations of \texttt{CoRe}$^2$ as a supplement to Fig.~\ref{fig:main_visualization} in the main paper. Specifically, for SD3.5, we showcase the synthesized images generated by \texttt{CoRe}$^2$ across different styles: anime, concept art, painting, and photography, which are displayed in Figs.~\ref{fig:anime-sd35},~\ref{fig:concept-sd35},~\ref{fig:paint-sd35} and~\ref{fig:photo-sd35}, respectively. For SDXL, we also showcase the synthesized images generated by \texttt{CoRe}$^2$ across different styles: anime, concept art, painting, and photography, which are displayed in Figs.~\ref{fig:anime-sdxl},~\ref{fig:concept-sdxl},~\ref{fig:paint-sdxl} and~\ref{fig:photo-sdxl}, respectively.

\begin{figure*}
    \centering
    \includegraphics[width=.9\linewidth]{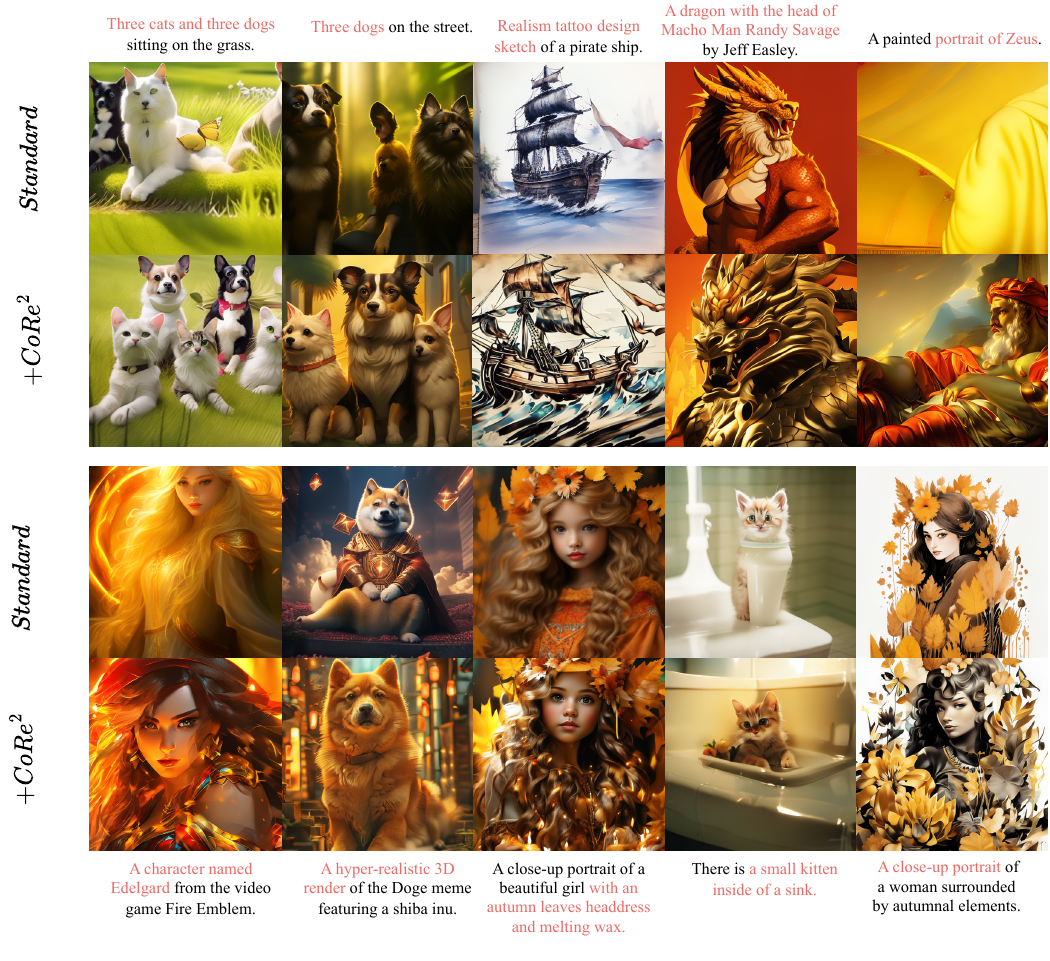}
    \vspace{-10pt}\caption{Visualization comparison about the standard sampling and \texttt{CoRe}$^2$ on LlamaGen.}
    \label{fig:llamagen}
    \vspace{-10pt}
\end{figure*}

%% file: tables/apd_flux_pick.tex
\begin{table*}[!t]
\centering
\vskip -0.01in
\small
\caption{Comparison of FLUX on Pick-of-Pic with different W2S guidance scale $\omega_\textrm{w2s}$.}
\scalebox{0.8}{
\setlength{\tabcolsep}{12pt}
\begin{tabular}{lcccc}
\Xhline{3\arrayrulewidth}\\[-2ex]
{\textbf{$\omega_\textrm{w2s}$}} & {\textbf{HPSV2 (↑)}} & {\textbf{PickScore (↑)}} & {\textbf{AES (↑)}} & {\textbf{ImageReward (↑)}}\\
\Xhline{3\arrayrulewidth}
0.7     & 28.71       & 22.13       & 6.05      & \CC{1.07} \\
0.85    & \CC{28.74}  & \CC{22.22}  & 6.07      & 1.02      \\
0.925   & 28.73       & 22.14       & \CC{6.10} & 1.01      \\
0.9625  & 28.67       & 22.11       & 6.09      & 0.98      \\
1.0     & 28.64       & 22.10       & 6.09      & 0.97      \\
\Xhline{3\arrayrulewidth}\\[-2ex]
\end{tabular}
}
\label{tab:pic_of_pick}
\vskip -0.07in
\end{table*}

%% file: tables/apd_flux_draw.tex
\begin{table*}[!t]
\centering
\vskip -0.01in
\small
\caption{Comparison of FLUX DrawBench between \texttt{CoRe}$^2$ and the standard sampling.}
\scalebox{0.8}{
\setlength{\tabcolsep}{12pt}
\begin{tabular}{lcccc}
\Xhline{3\arrayrulewidth}\\[-2ex]
{\textbf{Method}} & {\textbf{HPSV2 (↑)}} & {\textbf{PickScore (↑)}} & {\textbf{AES (↑)}} & {\textbf{ImageReward (↑)}}\\
\Xhline{3\arrayrulewidth}
Standard          & 29.75       & 22.85        & 5.82      & 0.98      \\
\texttt{CoRe}$^2$ (Ours) & \CC{29.76} & \CC{22.88}  & \CC{5.82} & \CC{0.98}      \\
\Xhline{3\arrayrulewidth}\\[-2ex]
\end{tabular}
}
\label{tab:drawbench}
\vskip -0.07in
\end{table*}

%% file: tables/apd_flux_geneval.tex
\begin{table*}[!t]
\centering
\vskip -0.01in
\small
\caption{Comparison of FLUX between \texttt{CoRe}$^2$ and the standard sampling on GenEval.}
\scalebox{0.8}{
\setlength{\tabcolsep}{12pt}
\begin{tabular}{lccccccc}
\Xhline{3\arrayrulewidth}\\[-2ex]
{\textbf{Method}} & {\textbf{Single (↑)}} & {\textbf{Two (↑)}} & {\textbf{Counting (↑)}} & {\textbf{Colors (↑)}} & {\textbf{Positions (↑)}} & {\textbf{Color Attribution (↑)}} & {\textbf{Overall (↑)}}\\
\Xhline{3\arrayrulewidth}
Standard          & 96.25\%     & 83.84\%     & 68.75\%    & \CC{79.79\%}   & 79.79\%   & 48.00\%      & 66.93\%     \\
\texttt{CoRe}$^2$ (Ours) & \CC{97.50\%} & \CC{87.88\%} & \CC{72.50\%} & 76.60\%   & \CC{24.00\%} & \CC{44.00\%} & \CC{67.08\%} \\
\Xhline{3\arrayrulewidth}\\[-2ex]
\end{tabular}
}
\label{tab:geneval}
\vskip -0.07in
\end{table*}

%% file: tables/apd_flux_hpd.tex
\begin{table*}[!t]
\centering
\vskip -0.01in
\small
\caption{Comparison of FLUX on HPD v2 between \texttt{CoRe}$^2$ and the standard sampling.}
\scalebox{0.8}{
\setlength{\tabcolsep}{12pt}
\begin{tabular}{lcccc}
\Xhline{3\arrayrulewidth}\\[-2ex]
{\textbf{Method}} & {\textbf{HPSV2 (↑)}} & {\textbf{PickScore (↑)}} & {\textbf{AES (↑)}} & {\textbf{ImageReward (↑)}}\\
\Xhline{3\arrayrulewidth}
Standard          & 29.08       & 22.70       & 6.01      & 1.05      \\
\texttt{CoRe}$^2$ (Ours) & \CC{29.15} & \CC{22.78} & \CC{6.01}      & \CC{1.07} \\
\Xhline{3\arrayrulewidth}\\[-2ex]
\end{tabular}
}
\label{tab:hpdv2}
\vskip -0.07in
\end{table*}